\documentclass{adobe_research}
\usepackage[T1]{fontenc}
\usepackage{lmodern}
\usepackage{multirow}
\usepackage{soul}
\usepackage{pifont}
\usepackage[T1]{fontenc}

\usepackage{graphicx}
\usepackage{amsmath,amssymb}

\usepackage{makecell}
\usepackage[dvipsnames]{xcolor}
\usepackage{color, colortbl}
\usepackage{caption}
\usepackage{algorithm}

\newcommand{\Position}{\mathbf{P}}

\newcommand{\Action}{\mathcal{A}}
\newcommand{\Frame}{{f}}
\newcommand{\Text}{\text{c}_{\text{text}}}

\usepackage{xspace}
\makeatletter
\DeclareRobustCommand\onedot{\futurelet\@let@token\@onedot}
\def\@onedot{\ifx\@let@token.\else.\null\fi\xspace}

\def\eg{\emph{e.g}\onedot} 
\def\ie{\emph{i.e}\onedot}

\makeatother

\usepackage{newtxtext}

\usepackage{algpseudocode}
\definecolor{catgray}{gray}{0.92}

\usepackage[dvipsnames]{xcolor} 
\definecolor{FutureOrange}{HTML}{EC866D}
\newcommand{\ours}{\textcolor{FutureOrange}{\textbf{RELIC}}}

\title{RELIC: Interactive Video World Model\\with Long-Horizon Memory}

\author[*\dagger]{Yicong Hong}
\author[*]{Yiqun Mei}
\author[*]{Chongjian Ge}
\author[]{\\Yiran Xu}
\author[]{Yang Zhou}
\author[]{Sai Bi}
\author[]{Yannick Hold-Geoffroy}
\author[]{Mike Roberts}
\author[]{Matthew Fisher}
\author[]{\\Eli Shechtman}
\author[]{Kalyan Sunkavalli}
\author[]{Feng Liu}
\author[]{Zhengqi Li}
\author[]{Hao Tan}

\contribution[*]{First Authors in Random Order}
\contribution[\dagger]{Project Lead.}

\abstract{
A truly interactive world model requires three key ingredients: real-time long-horizon streaming, consistent spatial memory, and precise user control. However, most existing approaches address only one of these aspects in isolation, as achieving all three simultaneously is highly challenging—for example, long-term memory mechanisms often degrade real-time performance. In this work, we present \ours, a unified framework that tackles these three challenges altogether. Given a single image and a text description, \ours\ enables memory-aware, long-duration exploration of arbitrary scenes in real time.
Built upon recent autoregressive video-diffusion distillation techniques, our model represents long-horizon memory using highly compressed historical latent tokens encoded with both relative actions and absolute camera poses within the KV cache. This compact, camera-aware memory structure supports implicit 3D-consistent content retrieval and enforces long-term coherence with minimal computational overhead. In parallel, we fine-tune a bidirectional teacher video model to generate sequences beyond its original 5-second training horizon, and transform it into a causal student generator using a new memory-efficient self-forcing paradigm that enables full-context distillation over long-duration teacher as well as long student self-rollouts.
Implemented as a 14B-parameter model and trained on a curated Unreal Engine–rendered dataset, \ours\ achieves real-time generation at 16 FPS while demonstrating more accurate action following, more stable long-horizon streaming, and more robust spatial-memory retrieval compared with prior work. These capabilities establish \ours\ as a strong foundation for the next generation of interactive world modeling.

}

\date{December 1st, 2025}
\adobedata[Project Page]{\url{https://relic-worldmodel.github.io/}}

\begin{document}

\maketitle

\section{Introduction}
\label{sec:introduction}

\begin{figure}[!h]
    \centering
    \vspace{-6mm}
    \includegraphics[width=1\textwidth]{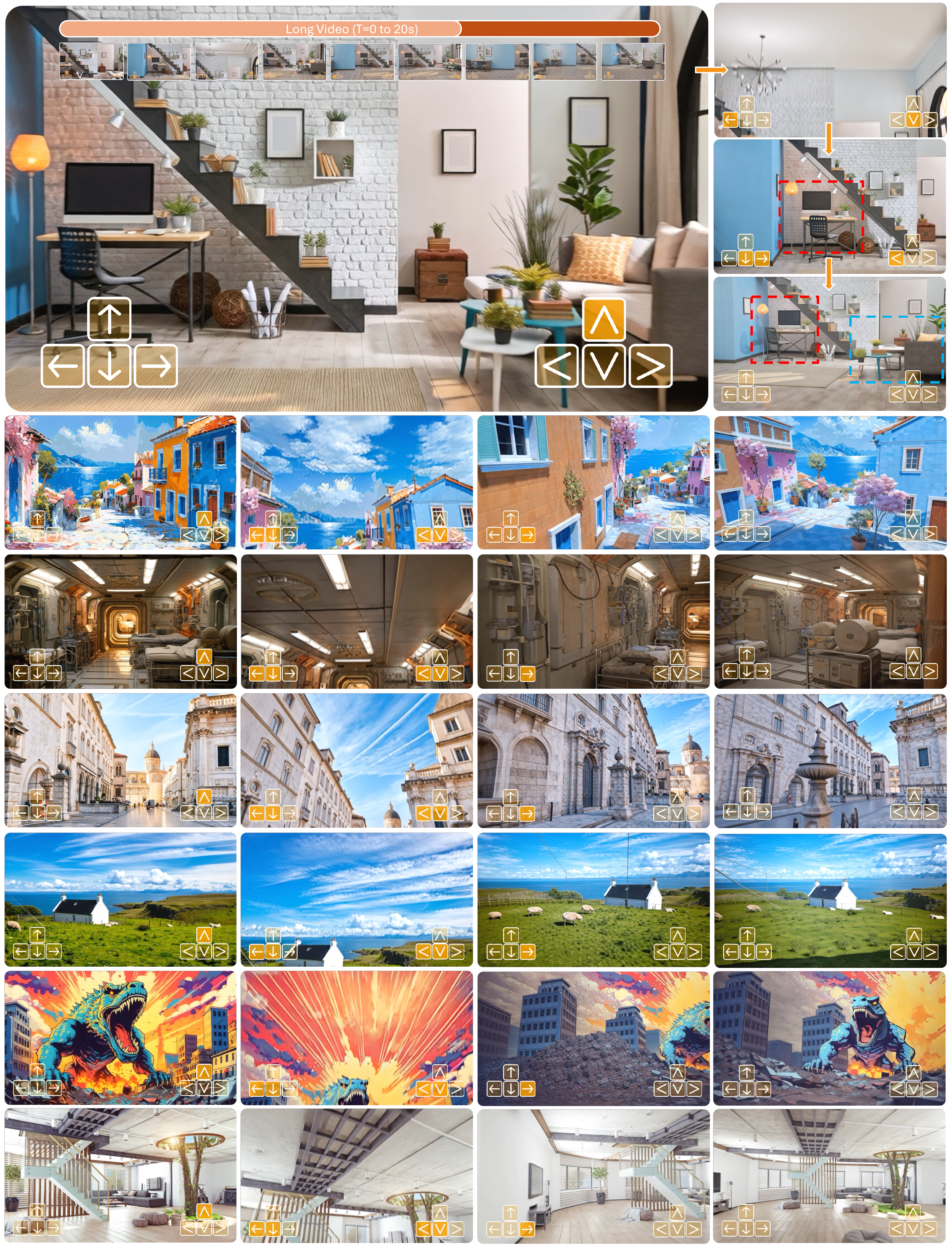}
    \vspace{-6mm}
    \caption{
        \ours\ is an interactive video world model that allows users to freely explore virtual scenes initialized from an arbitrary first-frame image in real time. Built as a 14B-parameter autoregressive model, \ours~generates videos at 480×832 resolution, 16 FPS, for up to 20 seconds, exhibiting consistent long-horizon spatial memory.
    }
    \label{fig:teaser}
    \vspace{-4mm}
\end{figure}

Imagine stepping into a picture and freely exploring or interacting with the world behind it in real time. \textit{World modeling}, which embodies this overarching vision, has gained significant attention for its potential to understand and simulate our three-dimensional physical world~\citep{ha2018worldmodels,worldlabs2025,videoworldsimulators2024}. By enabling interaction between an agent and its surrounding environment, world models can create a wide variety of real-world scenarios, facilitating downstream applications such as autonomous driving~\citep{hu2023gaia,tu2025drivingworldmodel,zhou2025hermes}, spatial intelligence~\citep{parkerholder2024genie2,genie3}, and robotics~\citep{bardes2023v,roth2025learned,gu2025cosmos}.
Recent advances in video generation thorough diffusion models~\citep{wan2025wan,gao2025seedance,lin2025apt2,wiedemer2025veo,polyak2024moviegen,kong2024hunyuanvideo} have demonstrated compelling potential to realize such immersive and interactive experiences by learning controllable autoregressive (AR) video models from large-scale synthetic and real-world datasets~\citep{li2025hunyuangame,wu2025spatial,feng2024thematrix,zhang2025matrixgame1,he2025matrixgame2,ye2025yan,parkerholder2024genie2,genie3,team2025hunyuanworld,yu2025gamefactory,mao2025yume,huang2025vid2world,chen2025deepverse4d}.

There are two essential foundation for building a practical video world model: 1) \emph{real-time long-horizon streaming} and 2) \emph{consistent spatial memory}.  
Real-time long-horizon streaming requires that video be generated at real-time latency in response to a continuous stream of user controls, such as camera movements or keyboard inputs. Consistent spatial memory, on the other hand, ensures that the model does not forget previously seen or explored scene content. Together, these capabilities enable the model to iteratively construct and maintain a persistent underlying 3D world.
In the context of autoregressive (AR) video diffusion models, real-time latency and throughput demand few-step or even one-step denoising models~\citep{song2023consistency,yin2024dmd1,yin2024dmd2,yin2025causvid, lin2025apt2, wang2025error}. Long-horizon streaming additionally requires stable autoregressive rollouts without drift~\citep{xie2025progressive,ruhe2024rollingdiff,chen2025skyreels2,sun2025ardiffuse,kodaira2025streamdit, huang2025self,liu2025rollingforcing,cui2025self,yang2025longlive}. Achieving consistent memory further requires storing long histories of past generated information—either within the KV cache or through dedicated memory mechanisms~\citep{yu2025contextasmem,xiao2025worldmem,ma2025yousee,ren2025gen3c,yu2025wonderworld,huang2025voyager}.
However, satisfying these requirements simultaneously is challenging: maintaining long-term spatial consistency typically incurs substantially higher computational complexity and memory bandwidth bottleneck, which directly conflicts with the need for real-time responsiveness.

\begin{table*}[t]
  \caption{\textbf{Comparison with recent video world models.} Note that generalization and memory often trade off against output duration and dynamic level. In particular, most of the models that can produce infinite-length stable videos in training-aligned domains do not generalize to arbitrary scenes or actions. Also, the generation speed largely depends on the model size, GPU hardware, and the degree of parallelism. $\texttt{T}$, $\texttt{R}$, and $\texttt{E}$ are translation-, rotation-, and other interaction-related actions, respectively.}
  \vspace{-5pt}
  \label{tab:main}
  \centering
  \resizebox{\textwidth}{!}{
  \begin{tabular}{lcccccccccc}
    \toprule
     & The Matrix & Genie-2 & GameCraft & Yume & Yan & Matrix-Game 2.0 & Genie-3 & \ours~(ours) \\
    \midrule
    \midrule
    Data Source & AAA Games & Unknown & AAA Games & Sekai & 3D game & Minecraft+UE+Sekai & Unknown & UE\\
    Action Space & $4\texttt{T}4\texttt{R}$ & $5\texttt{T}4\texttt{R}2\texttt{E}$ &  $4\texttt{T}4\texttt{R}$ & $4\texttt{T}4\texttt{R}$ & $7\texttt{T}2\texttt{R}$ & $4\texttt{T}$ & $5\texttt{T}4\texttt{R}1\texttt{E}$ & $6\texttt{T}6\texttt{R}$ \\
    Resolution & 720$\times$1280 & 720$\times$1280 & 720$\times$1280 & 544$\times$960 & 1080$\times$1920 & 352$\times$640 & 704$\times$1280 & 480$\times$832 \\
    Speed      & 8-16 FPS & Unknown & 24 FPS & 16 FPS & 60 FPS & 25 FPS & 24 FPS & 16 FPS \\
    Duration   & Infinite & 10-20 sec & 1 min & 20 sec & Infinite & 1 min & 1 min & 20 sec \\
    Generalization & \ding{73} & \ding{73}\ding{73} & \ding{73}\ding{73} & \ding{73}\ding{73} & \ding{73} & \ding{73}\ding{73} & \ding{73}\ding{73}\ding{73} & \ding{73}\ding{73}\ding{73} \\
    Memory  & None & \ding{73} & \ding{73} & None & None & None & \ding{73}\ding{73}\ding{73} & \ding{73}\ding{73}\ding{73}\\
    Model Size & 2.7B & Unknown & 13B & 14B & Unknown & 1.3B & Unknown & 14B \\
    \bottomrule
  \end{tabular}} \\
  \vspace{-10pt}
\end{table*}

To tackle these fundamental challenges holistically, we present \ours, a 14B action-controlled interactive video world model that supports both long-horizon streaming and efficient long-term memory retrieval. Our framework builds upon the Self-Forcing paradigm~\citep{yin2025causvid,huang2025self}, which distills an autoregressive (AR) student video model from a bidirectional teacher using the Distribution Matching Distillation (DMD) technique~\citep{yin2024dmd1,yin2024dmd2}. Specifically, 
\begin{enumerate}
    \item To unlock efficient long-horizon streaming and robust spatial memory retrieval, we represent the autoregressive model’s memory as a set of highly compressed historical latents, in a similar spirit to~\citep{zhang2025framepack}, which are encoded with both relative and absolute camera-pose and stored within the KV cache. This design enables implicit 3D scene-content retrieval through viewpoint-aware context alignment, while the high compression ratio allows \ours{} to retain the entire memory history with high computational efficiency.
    Our method stands in contrast to approaches that maintain spatial memory through recurrent model updates~\citep{zhang2025lact,po2025long,dalal2025one}, which are fundamentally constrained by the capacity of the internal model state and often tailored to specific visual domains. It also differs from methods that introduce external memory banks with handcrafted retrieval heuristics~\citep{yu2025contextasmem,xiao2025worldmem}, or that integrate explicit 3D scene representations~\citep{ma2025yousee,ren2025gen3c,yu2025wonderworld,huang2025voyager}, which can introduce strong inductive biases and are frequently bottlenecked by reconstruction accuracy and runtime cost.
    \item To overcome the limitations of the 5-second short-context training window used in most prior work~\citep{lin2025apt2,yang2025longlive}, where a short teacher fails to provide long-horizon consistency or handle drastic viewpoint changes, both of which are essential for world modeling, we fine-tune the teacher model to generate 20-second sequences. This extended temporal horizon enables supervision that enforces spatial and temporal consistency over significantly longer trajectories.
    However, performing DMD over the full 20-second student rollout during distillation is computationally intractable. To address this, we introduce a replayed back-propagation technique that enables memory-efficient differentiation of student parameters by accumulating cached gradients of the DMD loss in a temporal block-wise fashion over the entire self-rollout.
\end{enumerate}

We curated 350 licensed Unreal Engine–rendered scenes, encompassing approximately 1600 minutes of training video with high-quality text, action, and camera-trajectory annotations. Importantly, we collect trajectories that include both single and mixed actions, as well as trajectories with viewpoint revisitations, enabling the model to learn precise decoupled controls, flexible action composition, and long-range memory retrieval. Overall, our architectural design enables both efficient training and inference: \ours\ achieves 16 FPS generation throughput on 4 H100 GPUs while maintaining precise action and text following, as well as long-horizon spatial consistency. 
We believe that the \ours\ framework provides an effective and flexible foundation for advancing interactive world modeling, paving the way for future capabilities across numerous domains.

\section{Related Works} 
\label{sec:relatedwork}

\paragraph{\textbf{Video World Models.}}
Building interactive video world models requires the integration of diverse techniques, spanning autoregressive video generation~\citep{yin2025causvid,huang2025self,teng2025magi,xie2025progressive,henschel2025streamingt2v,chen2025skyreels2,yang2025longlive,kodaira2025streamdit} and few-step model distillation~\citep{yin2024dmd1,yin2024dmd2,song2023consistency,geng2025meanflow}, alongside breakthroughs in several fundamental challenges, including long-context memory~\citep{po2025long,song2025history,zhang2025lact}, spatial consistency~\citep{xiao2025worldmem,huang2025voyager,zhang2025world,wu2025spatial,wu2025geometry,yu2025contextasmem}, and the generation of interactive entities~\citep{lu2025dreamart,xia2025drawer,kreber2025guiding,luo2025physpart,kurai2025magiccraft}. Recent industrial systems such as Matrix-Game~2.0~\citep{he2025matrixgame2}, Magica~2~\citep{magica2}, RTFM~\citep{rtfm2025}, and Genie-3~\citep{genie3} have demonstrated remarkable progress toward this goal, yet substantial advances are still required to make world models truly practical.

\paragraph{\textbf{Long Video Generation.}}
The \textit{drifting} phenomenon is a typical issue in long-horizon autoregressive video generation, manifesting as rapid degradation in video quality, color over-saturation, or even static outputs.
Early training-free approaches, such as FIFO-Diffusion~\citep{kim2024fifo}, introduce an inference strategy that operates on a window of latent frames with monotonically increasing noise levels.
After a fixed number of denoising steps, a clean latent is popped out (and cached), while a new noise sample is pushed in at the other end, enabling continuous video generation.
Following this idea and to bridge the train–test discrepancy, PA-VDM~\citep{xie2025progressive}, Rolling Diffusion~\citep{ruhe2024rollingdiff}, SkyReels-v2~\citep{chen2025skyreels2}, AR-Diffusion~\citep{sun2025ardiffuse}, StreamDiT~\citep{kodaira2025streamdit}, Rolling Forcing~\citep{liu2025rollingforcing}, and Wan’s Streamer~\citep{wan2025wan} incorporate the same noise-scheduling patterns during training, thereby achieving stable, minute-long video generation.
These scheduling strategies can be regarded as special cases of Diffusion-Forcing~\citep{chen2024dforcing}, where latents within the modeling window may carry uneven noise levels.
A key strength of this framework lies in its flexibility to incorporate context tokens during both training and inference~\citep{song2025history}, effectively serving as a form of memory, an essential component for consistent world simulation. Meanwhile, \textit{Teacher-forcing}, inspired by the next-token prediction paradigm of large language models, trains the model to predict new frames conditioned on previously generated and cached outputs~\citep{alonso2024atari,jin2024pyramidal,valevski2024gamengen,zhou2025tamingtf,hu2024acdit,gao2024ca2}.
However, such methods often suffer from severe \textit{drifting} only after a small number of rollouts.
To mitigate this issue, Self-Forcing~\citep{huang2025self} introduces self-rolling mechanisms that allow the model to adapt to its own predictions, analogous to \textit{student-forcing} in embodied navigation, where an agent learns to correct itself when deviating from a target trajectory~\citep{anderson2018vln,hong2020vlnbert}. Furthermore, APT-2~\citep{lin2025apt2} and LongLive~\citep{yang2025longlive} extend self-rolling training to minute lengths, enabling long-horizon generation.

\paragraph{\textbf{Data for World Exploration.}}
Building a world exploration model fundamentally requires high-quality data that are dynamic, long-horizon (spanning minutes), and paired with accurate action annotations, a combination that is rare on the web and expensive to obtain~\citep{li2025sekai,wang2025spatialvid,che2024gamegenx,feng2024thematrix,he2025matrixgame2}. Large-scale datasets such as Sekai~\citep{li2025sekai} offer abundant real-world walking and drone videos, but their action distributions are highly imbalanced and often compositional, which makes precise control learning difficult. Although AAA game datasets provide clean and reliable action–video pairs~\citep{feng2024thematrix,li2025hunyuangame,he2025matrixgame2}, their strong stylistic bias and limited scene diversity mean that models trained on such narrow-domain data, including specific game environments~\citep{ye2025yan,zhang2025matrixgame1} or videos with repetitive camera and scene patterns~\citep{zhang2025framepack,henschel2025streamingt2v}, often produce stable long-horizon rollouts but fail to generalize to broader and more dynamic environments. In this work, we observe that when starting from a strong pretrained backbone, a small amount of carefully curated, control-precise data collected in Unreal Engine is sufficient to improve controllability while preserving generalization capability.

\section{Data Curation for Interactive Video World Model}  \label{sec:data}

\subsection{Data Overview}
\begin{figure}[t]
    \centering
    \begin{subfigure}{0.49\linewidth}
        \centering
        \includegraphics[width=\linewidth]{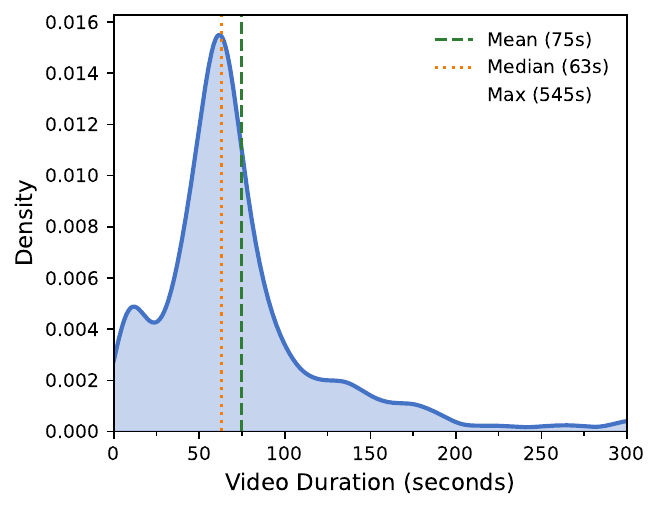}
        \caption{Video duration distribution of our curated UE dataset.}
        \label{fig:video_duration_distribution}
    \end{subfigure}
    \hfill
    \begin{subfigure}{0.49\linewidth}
        \centering
        \includegraphics[width=\linewidth]{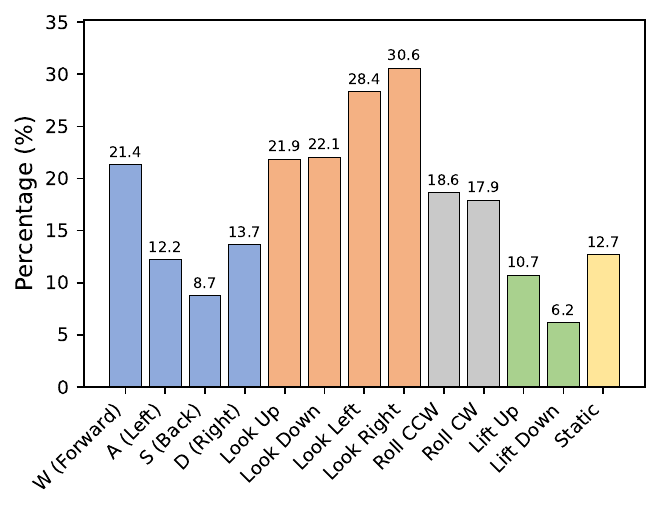}
        \caption{Action distribution of our curated UE dataset }
        \label{fig:action_distribution}
    \end{subfigure}
    \caption{\textbf{Dataset statistics visualization.} Left: video duration distribution; Right: action distribution.}
    \label{fig:dataset_stats}
\end{figure}

To support training a real-time, long-horizon interactive video world model, we construct a large-scale synthetic video dataset rendered entirely in Unreal Engine (UE). The dataset is designed to provide (i) diverse and complex 3D environments and (ii) precise control over camera trajectory.
Our final curated dataset contains over {1400} human-controlled camera trajectories collected from 350 high-quality licensed 3D scenes spanning both indoor (\eg, homes, offices) and outdoor environments (\eg, forests, mountains, streets). After applying the filtering procedures described in \cref{sec:data_filtering}, we obtain more than \textbf{1600 minutes} of high-fidelity 720p video sequences 
The clip duration distribution is shown in \cref{fig:video_duration_distribution}, with an average of $\sim$75 seconds and a maximum of up to 9 minutes.
We derive action labels from the corresponding camera motions (\cref{sec:data_annotation}), and sample them so that distribution of each action is well-balanced for training model.
We now summarize the data curation pipeline, filtering strategy, and annotation process below.

\subsection{Data Processing and Filtering} \label{sec:data_filtering}

We begin by curating 350 photorealistic static UE scenes covering a wide range of indoor and outdoor layouts. Human operators navigate each scene using a collision-constrained camera controller to ensure physically plausible movement. During navigation, we record continuous 6-DoF camera trajectories—including positions, orientations, and the corresponding timestamps, which are then rendered into high-quality 720p video sequences using the UE renderer.

This synthetic capture pipeline is designed to address the fundamental limitations of existing real-world navigation datasets. Prior work has primarily trained models on real-world video corpora, such as FOV walking datasets and drone-based navigation videos~\citep{wang2025spatialvid,li2025sekai}. However, these sources suffer from three key limitations.
(1) \textbf{Imbalanced action distributions}: real-world videos are dominated by forward motion, with very limited lateral or rotational movement, which we observe prevents models from learning diverse egocentric movement behaviors.
(2) \textbf{Overly coupled action behavior}: real-world videos often contain tightly coupled actions, such as turning while moving forward, which makes it difficult for the model to learn disentangled single-action control.
(3) \textbf{Lack of revisitation of past viewpoints}: real-world videos rarely return to previously visited locations over long horizons, weakening the model’s ability to learn long-context spatial recall and memory-based reasoning or retrieval for consistent world generation.

By contrast, our UE-rendered camera trajectories are intentionally curated to be both well-balanced in sampled action space and designed to frequently revisit locations and scene content in long-horizon settings, directly addressing the aforementioned limitations. Moreover, we observe that raw UE-rendered videos may contain various types of artifacts, including (1) overly fast or jittery camera motion, (2) unstable or collision-prone trajectories, and (3) rendering issues such as over-exposure or missing textures. These observations motivate a new filtering pipeline described below:

\begin{figure}[t]
    \centering
    \includegraphics[width=1.0\textwidth]{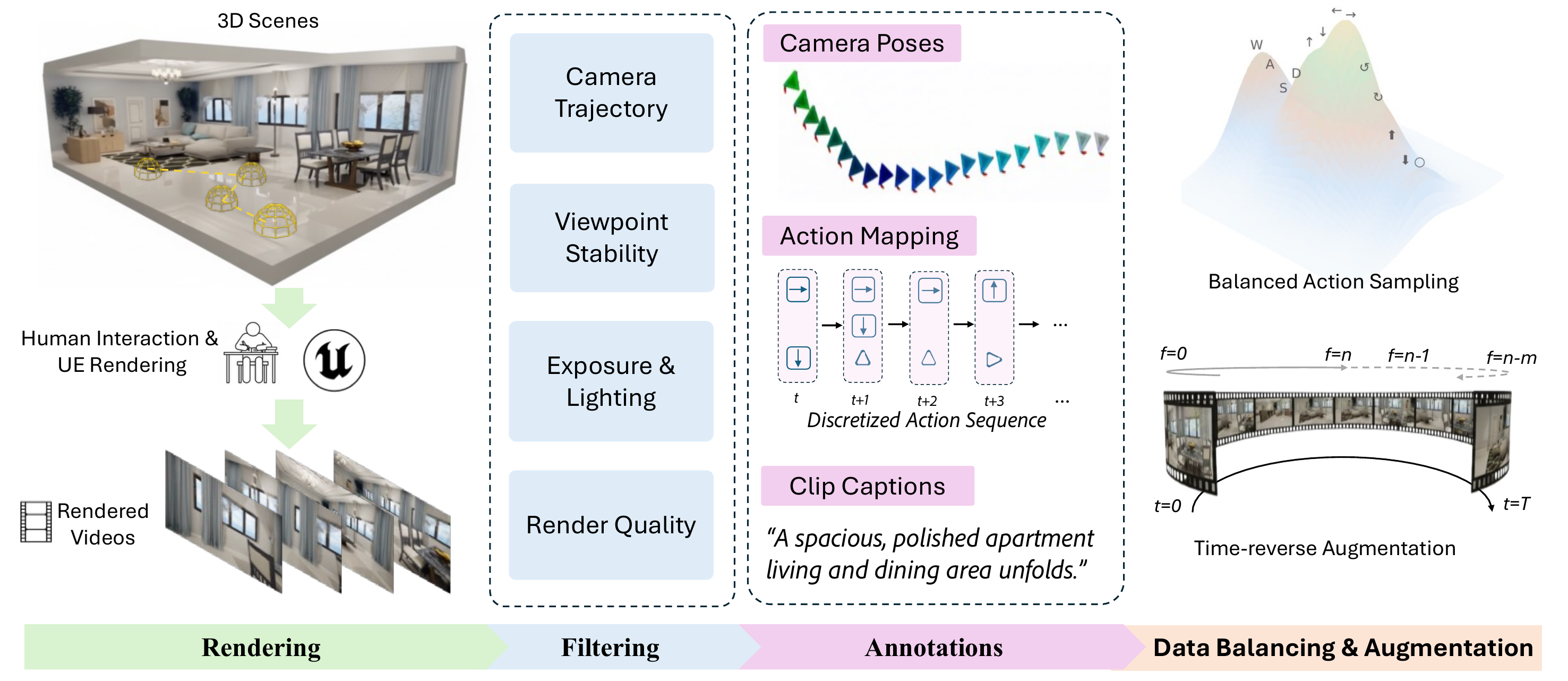}
    \caption{\textbf{The data curation pipeline in \ours.} Given a set of 3D scenes, we manually collect thousands of camera trajectories and generate high-quality video-action-text triplets through a series of data filtering, captioning, and balancing steps. }
    \label{fig:data_pipeline}
\end{figure}

\begin{itemize}
\item \textbf{Camera Motion.}
Trajectories exhibiting unnatural camera-motion patterns, such as excessive panning speed, abrupt rotations, or inconsistent velocity profiles introduced by human operators, are manually removed. This ensures that the camera dynamics remain smooth and physically plausible, preventing the model from learning unrealistic motion behavior.

\item \textbf{Viewpoint Stability.}
Segments exhibiting micro-jitters, oscillatory drift, or near-collision paths are excluded. Filtering out these unstable videos prevents high-frequency noise from contaminating the training signal.

\item \textbf{Exposure and Lighting Problems.}
Each clip is reviewed for exposure-related anomalies, including overexposed highlights, under-lit interiors, or illumination flicker caused by renderer settings or inconsistencies in scene assets.

\item \textbf{Render Quality.}
Clips with missing textures, geometry popping, incomplete meshes, or other rendering defects are discarded. Ensuring high-fidelity rendering is essential for high-quality world generation and video synthesis.
\end{itemize}

\subsection{Data Annotations} \label{sec:data_annotation}

\paragraph{\textbf{Camera Pose Annotations.}} 
For every rendered video clip $(\Frame_1, \Frame_2, ..., \Frame_{T})$, we record precise, frame-aligned camera pose annotations directly derived from the 6-DoF trajectories provided by Unreal Engine (UE). These camera annotations include full 6-DoF camera poses, which consist of absolute camera positions $(\Position_1, \Position_2, ... \Position_T)$  and world-to-camera camera orientations $(R_1, R_2, ..., R_T)$.
These complete camera annotations are essential ingredients for training interactive video world models because they allow accurate mappings between input action controls and the corresponding dynamic responses from the agent and environment. Furthermore, full 6-DoF camera poses also support effective long-context spatial content retrieval, as shown in the experiment section.

\paragraph{\textbf{Action Annotations.}} 
Although Unreal Engine records continuous 6-DoF camera trajectories, the control interface of our world model is typically structured around user actions, inputs that real users can realistically provide (\eg, moving forward, panning, or looking up), rather than continuous camera poses.
To bridge this gap, we convert continuous 6-DoF camera-pose sequences into per-frame action labels $\Action \in \mathbb{R}^{13}$, \ie, 13-DOF input-control format described in~\cref{subsec:action_control}. Our action-annotation pipeline derives per-frame actions $\Action_{t}$ by comparing the relative camera poses between adjacent frames $(\Frame_t, \Frame_{t+1})$ in the video (see \cref{alg:ue_actions} for details).

\begin{algorithm}[t]
\caption{Action Label Extraction from UE Camera Trajectories}
\label{alg:ue_actions}
\begin{algorithmic}[1]

\Require Annotation JSON $\mathcal{J}$
\Ensure  Action sequence $\mathcal{A}_{1:T}$ (including, movement action $\mathbf{A}^\mathrm{m}$, and rotate action $\mathbf{A}^\mathrm{r}$)

\Function{ReadUEActions}{$\mathcal{J}$}
    \State Load frames $\{f_t\}_{t=1}^T$, extract $\mathbf{P}_t$ and $R_t$
    \State $\Delta \mathbf{P}_t^{\mathrm{c}} \gets R_t(\mathbf{P}_{t+1} - \mathbf{P}_t)$
    \State $\bar{d} \gets$ mean movement magnitude over non-zero $\Delta \mathbf{P}_t^{\mathrm{c}}$
    \For{$t=1$ to $T-1$}
        \State $\mathbf{A}_t \gets \Call{InferPairAction}{\Delta \mathbf{P}_t^{\mathrm{c}}, \bar{d}, t}$
    \EndFor
    \State Prepend static action; \Return $\mathbf{A}_{1:T}$
\EndFunction

\vspace{3pt}
\Function{InferPairAction}{$\Delta \mathbf{P}_t^{\mathrm{c}}, \ \bar{d},\ t$}
    \State // \textbf{Translational motion in camera coordinates}
    \State $(d_f, d_s, d_z) \gets \Delta \mathbf{P}^{\mathrm{c}}_t / \bar{d}$
    \State Assign $\mathbf{A}^\mathrm{m}:\{\texttt{w/s},\texttt{a/d},\texttt{lifting\_up/down}\}$ from $d_f,d_s,d_z$
    
    \State // \textbf{Camera rotation}
    \State $\Delta R^c_{t} \gets R_{t+1}(R_t)^\top$
    \State $(\Delta\text{yaw},\Delta\text{pitch},\Delta\text{roll}) \gets$ Euler decomposition of $R_{\text{rel}}$
    \State Assign $\mathbf{A}^\mathrm{r}:\{\texttt{camera\_up/down},\texttt{left/right},\texttt{roll\_ccw/cw}\}$ from the angles
    \State Mark \texttt{static} if no action activated
    \State \Return action vector $(\mathbf{A}^m, \mathbf{A}^r)$
\EndFunction

\end{algorithmic}
\end{algorithm}

Specifically, we first compute the relative camera-translation vector $\Delta \Position^c_t$ in the camera’s egocentric coordinate system using the camera positions in world coordinates $\Position_t$ and the world-to-camera rotation matrix $R_t$ at time $t$:
\begin{equation}
\Delta \Position^c_t = R_{t} (\Position_{t+1} - \Position_t).
\end{equation}
The $x$, $y$, and $z$ components of $\Delta \Position^c_t$ constitute the translational 3-DOF component of our action labels. This vector represents the per-frame camera displacement, \ie, the instantaneous motion between two consecutive frames.
Importantly, we observe that different 3D scenes may exhibit different inherent scales, analogous to the scale ambiguity discussed in the Structure-from-Motion (SfM) literature. To ensure consistent behavior across environments, we normalize each 3D displacement vector by the average displacement magnitude of the entire clip (computed across all consecutive frame pairs), which characterizes the typical motion scale of that trajectory.
This normalization interprets each $\Delta \Position^c_t$ as a relative motion ratio, indicating how the current displacement compares to the typical displacement within the clip. During inference, the average displacement magnitude (a coefficient $\gamma$) can be adjusted to control the overall motion scale of the generated videos.

Similarly, we annotate 6-DoF rotational actions by computing the relative camera rotation $\Delta R^c_t$ between adjacent frames:
\begin{equation}
\Delta R^c_t = R_{t+1} (R_t)^{{T}},
\end{equation}
% where $R_t$ denotes the world-to-camera rotation matrix at time $t$. 
We then convert $\Delta R^c_t$ into yaw, pitch, and roll Euler angles following UE’s intrinsic rotation convention. These three angles constitute the rotational portion of the action labels.
Moreover, if all translational and rotational components of a relative pose are near zero, we assign the action label at the current time step to $\texttt{static}$.

Each frame is ultimately assigned a 13-dimensional action vector that mirrors the model’s input-control interface. This structured representation enables the model to learn a consistent mapping from per-timestep discrete controls to the corresponding dynamic responses of the agent and environment. To improve action-following capability across a wide range of user-input scenarios and to reduce distributional bias during long-horizon generation, we intentionally collect a diverse and well-balanced set of actions during data acquisition, unlike many real-world datasets~\citep{li2025sekai,mao2025yume,wang2025spatialvid} that are dominated by forward or dolly-in motions.

\paragraph{\textbf{Segmented Captions}} 
Long videos in our UE dataset can span multiple minutes, making a single global text caption insufficient to describe the content of the entire video. To reduce visual–caption misalignment, we split each long video into 5-second segments and generate a short, high-level caption for each segment using GPT-5~\citep{openai_gpt5_2024}. During training, when the sampled video length exceeds 5 seconds, we use the text description corresponding to the first segment within the sampled video as the caption prompt for that training sample.

Because text descriptions can sometimes contradict user-intended action inputs—for example, a user may wish to move forward in the environment while the text inadvertently describes the camera moving backward or panning—we thus attempt to avoid any contradiction between text conditioning and user-input actions at inference time. To achieve this, captions are explicitly constrained to describe only static, scene-level attributes. Specifically, we prompt GPT-5 to suppress descriptions of camera motion and fine-grained object motion, as we empirically found that common vision–language models~\citep{bai2023qwen, yuan2025tarsier2, chen2024internvl} tend to overemphasize such dynamics when captioning a video. We observe that this strategy enables more precise joint action and text control for AR video generation.

\subsection{Data Augmentation}
Although our curated camera trajectories deliberately encourage revisitation of past viewpoints, random segment sampling during training does not guarantee that most sampled sequences contain sufficient signal for the model to learn long-context memory retrieval. Therefore, to strengthen the model’s spatial long-context modeling capability, we propose a simple yet effective time-reverse augmentation technique that injects controlled time-reversal structure into the training clips.

Specifically, as shown in \cref{fig:data_pipeline} given a sampled training video segment of length $T$, we uniformly sample a pivot index $t^\ast$ from the second half of the clip, $t^\ast \sim \mathcal{U}(T/2,, T)$, and construct a palindrome-style training sequence by concatenating the forward segment $f_{1:t^\ast}$ with its time-reversed counterpart $f_{t^\ast:(2t^\ast - T)}$. This augmentation produces a clip in which the visual content naturally “looks back’’ in time, encouraging the model to learn effective spatial recall and to leverage long-horizon memory.

\section{\ours\ World Model}

Our goal is to generate a video stream
$(\hat{\Frame_1}, \hat{\Frame_2}, \ldots, \hat{\Frame_T})$
from an input RGB image ${\Frame_0}$ and a text description $\Text$, given a stream of action-control inputs (\eg, keyboard or mouse commands)
$(\Action_1, \Action_2, \ldots, \Action_T)$.
The model must preserve spatial and temporal consistency over long horizons while running in real time with interactive responsiveness to user inputs.

Following recent work~\citep{huang2025self, yin2025causvid, shin2025motionstream}, we propose a two-stage pipeline that distills a few-step autoregressive (AR) video diffusion model from a bidirectional video diffusion teacher model, where both models are conditioned on the additional action-control signal. However, unlike prior work, which typically focuses either on real-time performance~\citep{lin2025apt2} or on long-term generation~\citep{yang2025longlive, liu2025rolling}, achieving both capabilities simultaneously—along with long-horizon memory—is substantially more challenging, as these requirements often conflict. In particular, long-horizon spatial memory demands additional computation and GPU memory to store, transfer, and reason over past tokens, which introduces significant FLOPs and memory-bandwidth bottlenecks for real-time applications.

Our \ours\ model is designed to address these challenges by introducing several key innovations into the framework. First, we redesign the bidirectional video diffusion teacher model to support long-duration (20-second) generation with strong action-following ability (\cref{subsec:base_model}). To enable low-latency interactive streaming, we convert the teacher into an AR video-diffusion model and devise a spatial-aware memory mechanism that efficiently compresses historical video tokens within the KV cache. We further combine this with a training strategy that induces emergent memory retrieval from action–video data (\cref{sec:self_rollout}). Next, we introduce a new variant of the self-forcing training paradigm that enables memory-efficient distillation of a few-step AR model even under the teacher’s 20-second long context (\cref{subsec:self_forcing}). Lastly, we incorporate additional runtime optimizations that bring the model inference to real-time (\cref{subsec:runtime}).

\subsection{Action-Conditioned Teacher for Long Video Generation} 
\label{sec:method}

\subsubsection{Base Architecture} \label{subsec:base_model}

Our framework is built upon Wan-2.1~\citep{wan2025wan}, a bidirectional video diffusion model pretrained for for 5-second text-to-video and image-to-video generation. The base model consists primarily of a spatio-temporal variational autoencoder (ST-VAE) and diffusion transformers (DiTs). The ST-VAE employs a 3D causal architecture that maps between high-dimensional video space and a lower-dimensional latent token space, achieving an $8\times$ spatial compression ratio and a $4\times$ temporal compression ratio. Its temporally causal design, together with an internal feature-cache mechanism, plays a crucial role in enabling our real-time video streaming capabilities.

The video DiTs include patchifying and unpatchifying layers along with a series of transformer blocks. Input text descriptions are encoded with umT5~\citep{Chung2023UniMaxFA} and integrated into the model via cross-attention, while denoising timestep embeddings are injected through a shared MLP module. Our \ours\ adopts the 14B-parameter Wan-2.1 text-to-video DiT model due to its strong ability to understand 3D structure, generate consistent scene content under large camera motions, and retrieve long-term memory. Since our goal is to accept both text and an input image as conditioning signals, during both training and inference, we feed the clean latent corresponding to the input frame into the model and always set its noise level to zero when it is concatenated with other noisy video latents. 

To enable precise keyboard control, long-duration video streaming, and long-horizon spatial memory—the three key capabilities required for a video world model—we must first construct a robust action-conditioned \emph{teacher} model capable of high-quality long-form video generation from streaming action inputs. In the following section, we describe our design choices for integrating action-control signals into the base architecture and extending its generation horizon beyond the original 5-second limit.

\begin{figure}[t]
    \centering
    \includegraphics[width=1.0\textwidth]{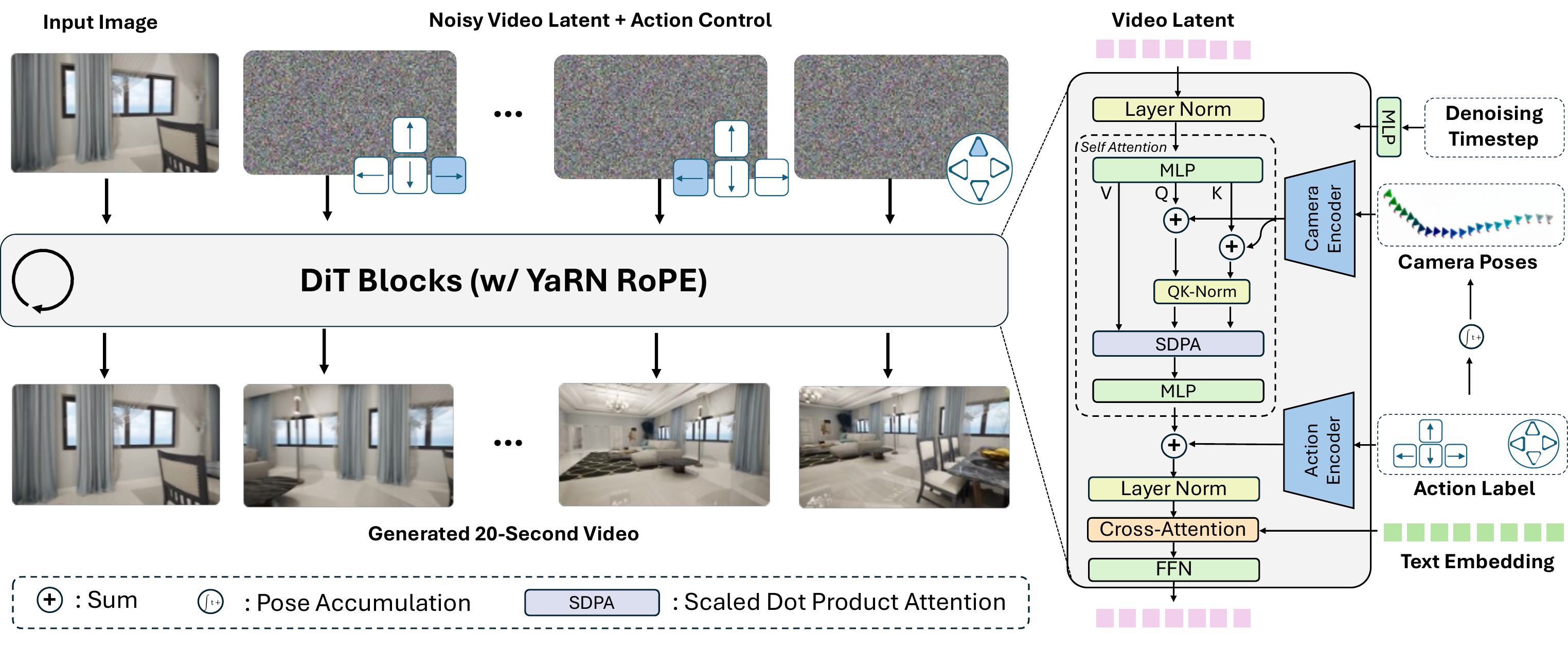}
    \caption{\textbf{Model Pipeline.} Starting from an input image and a sequence of noisy video latents, our DiT-based architecture generates a 20-second video conditioned on text, action labels, and camera poses. Each DiT block integrates YaRN-RoPE, SDPA with QK-Norm, and cross-attention to conditioning tokens. Camera and action information are embedded through dedicated encoders, and conditioning is injected throughout the denoising process to produce spatially consistent and action-aligned video frames. }
    \label{fig:model_pipeline}
\end{figure}

\subsubsection{Action Control}
\label{subsec:action_control}

\paragraph{\textbf{Action Space.}}
We design a 13-degree action space $\Action \in \mathbb{R}^{13}$ for \ours, enabling full 6-DoF camera viewpoint control. Specifically, $\Action$ specifies the magnitudes of six translational motions—\texttt{Dolly In $\uparrow$, Dolly Out $\downarrow$, Truck Left $\leftarrow$, Truck Right $\rightarrow$, Pedestal Up}, and \texttt{Pedestal Down}—and six rotational motions—\texttt{Tilt Up $\wedge$, Tilt Down $\vee$, Pan Left $<$, Pan Right $>$, Roll Clockwise}, and \texttt{Roll Counter-Clockwise}—between consecutive frames in the video, along with a static action \texttt{[Static]} representing no camera movement. 
Each action is represented as a non-negative scalar value rather than a binary flag, thus it can encode the relative translational and rotational velocities induced by user inputs.

\paragraph{\textbf{Action Conditioning .}}

To enhance action following and spatial memory retrieval, our \ours\ model incorporates not only the 13-DoF relative action controls but also the derived 6-DoF absolute camera poses as additional conditioning signals. Because the action vector $\Action \in \mathbb{R}^{13}$ represents relative camera motion—\ie, the translational velocity $\Delta \Position^c_t$ and rotational velocity $\Delta R^c_t$ between frames at times $t$ and $t+1$—we obtain the absolute camera poses $\Position_t \in \mathbb{R}^3$ and $R_t \in SO(3)$ by integrating the relative motions:

\begin{align}
    \Position_t = \sum_{i=1}^{t} (R_i)^T \Delta \Position^c_i , \quad R_t = \prod_{i=1}^{t} \Delta R_i^{c}
\end{align}

Using continuous-valued action encoding allows the model to represent motion strengths consistent with observed frame transitions, enabling it to accommodate videos captured under different unit velocities. During inference, we stream a continuous sequence of user actions to the model from keyboard inputs, represented as a multi-hot vector scaled by a predefined coefficient $\gamma$, which modulates the magnitude of camera motion in the generated video.

We embed both the relative actions $\Action$ and the absolute camera poses $(\Position_t, R_t)$ using two dedicated encoders and inject their embeddings into every transformer block through distinct pathways, as illustrated in~\cref{fig:model_pipeline}. In particular, each encoder consists of a temporal patchifying module followed by an MLP modulation layer that temporally compresses the control signal by $4\times$. Relative action embeddings are added directly to the latents after the self-attention layer, while absolute camera pose embeddings are added to the \textit{query} $(Q)$ and \textit{key} $(K)$ projections before the scaled dot-product attention (SDPA), with the \textit{value} $(V)$ projections left unchanged.
This design reflects the distinct computational roles of the two signals: relative actions guide the model in generating frame-to-frame scene transitions consistent with user control, whereas absolute camera poses act as proxies for retrieving spatial content across viewpoints and time.

\subsubsection{Long-Horizon Training} 
\label{subsec:base_train}
The original Wan-2.1 model is pretrained to generate 5-second videos (81 frames) at 16 FPS. Although recent work~\citep{lin2025apt2, yang2025longlive} has demonstrated that using a teacher model with a short context window can distill a student model capable of streaming long video sequences, we argue that a 5-second video training context length is typically not sufficient to enable the capability for long-term memory retrieval or be robust to significant camera viewpoint changes. The model must be trained directly under a long-horizon setting to learn how to restore spatial scene contents previously seen through the memory. Therefore, the first requirement is to enable the teacher model having the capacity for long-duration video generation, and we start to fine-tune our action-conditioned video diffusion model on our action-video dataset (\cref{sec:data}) to extend its generation duration to 20-seconds (317 frames). 
In particular, we employ a curriculum learning strategy: the model is first trained on 5-second videos for 5,000 iterations, followed by 10-second videos for 1,000 iterations, and finally by 20-second videos for another 4,000 iterations.
To facilitate more rapid adaptation to longer sequences, we apply the YaRN technique~\citep{peng2023yarn} to extend the Rotary Positional Embeddings (RoPE)~\citep{su2024rope} for the query and key tokens in each self-attention layer of the DiT block.

\subsection{Autoregressive Student for Real-Time Streaming} \label{sec:self_rollout}

Designing our final interactive video world model entails three core challenges: \textbf{(1) Memory}, \ie the ability to recall and faithfully re-render previously generated scenes when the camera agent revisits earlier viewpoints; \textbf{(2) Streaming inference}, to enable low-latency, real-time interactive exploration; and \textbf{(3) Long-horizon generation}, to provide sufficiently extended temporal context for user navigation and discovery.

To address these challenges, we distill our 20-second bidirectional teacher model into a few-step, memory-aware causal student video diffusion model. Our student model is also built upon Wan2.1-14B~\citep{wan2025wan}, but replaces bidirectional attention with block-wise causal attention and generates latent frames in a block-causal manner, following similar principles as recent autoregressive video diffusion approaches~\citep{yin2025causvid, huang2025self, yang2025longlive}. We describe our memory mechanisms and training strategies below, and refer readers to these prior works for the full mathematical formulation of autoregressive video generation.

\subsection{Memory} \label{subsec:memmory}

Most recent autoregressive (AR) video models adopt causal attention over a short sliding window to enable efficient long-video streaming \citep{yang2025longlive, shin2025motionstream}. While this strategy effectively reduces inference latency and improves throughput, it inherently limits the model’s ability to retrieve long-range information, which is crucial for consistent world modeling. A straightforward solution to restore such long-horizon consistency is to retain all past tokens in the KV cache during AR inference. However, both the KV-cache memory footprint and the per-token attention cost scale linearly with sequence length. Consequently, as video length increases during AR generation, both computation and communication overhead grow proportionally, making real-time streaming inference particularly challenging.

To address this dilemma, we introduce a memory-compression mechanism, illustrated in \cref{fig:dmd}. Given a newly denoised latent at index $i$, our KV cache is composed of two branches: a rolling-window cache and a compressed long-horizon spatial memory cache. The rolling-window cache stores uncompressed KV tokens for recent video latent between indices $i-w$ and $i$, where $w$ denotes the sliding-window size. We maintain a small rolling window to prevent the model from relying solely on short-term patterns and encourage effective learning from the compressed long-range memory.
The compressed long-horizon spatial memory cache, in contrast, stores spatially downsampled KV tokens for video latent from the beginning of the sequence up to index $i-w$, following a predefined compression schedule. In practice, we adopt an empirically balanced configuration that interleaves spatial downsampling factors of $1\times$ (no compression), $2\times$, and $4\times$ across latents.
This design is motivated by the observation that VAE-encoded latent spaces exhibit substantial spatial redundancy, allowing much of the original information to remain recoverable after moderate spatial compression. As a result, the model can still reconstruct high-fidelity content from the compressed long-range context. On average, our strategy reduces the total token count by approximately $4\times$ (\eg, from $120$K to $30$K), which in turn yields a proportional $4\times$ reduction in both KV-cache memory and attention FLOPs.

\begin{figure}[t!]
    \centering
    \includegraphics[width=\textwidth]{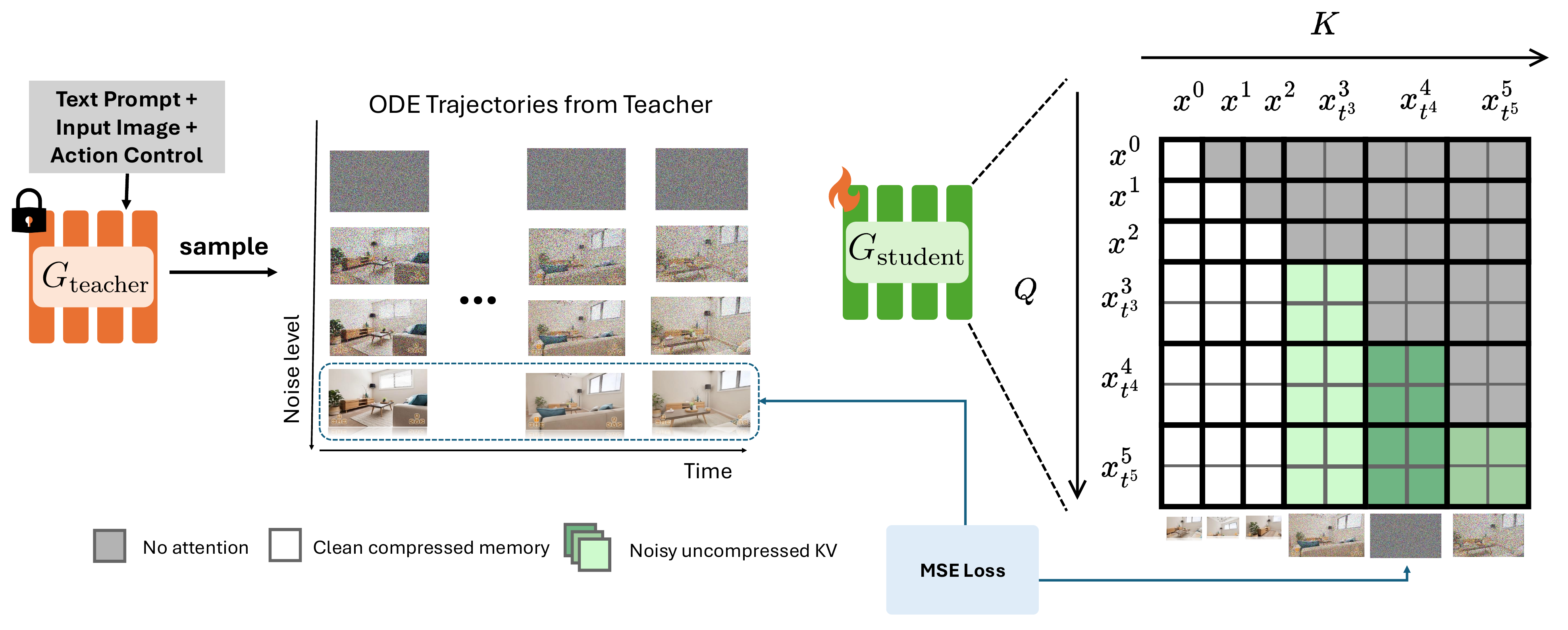}
    \caption{\textbf{ODE initialization.} We convert a bidirectional video diffusion model into a causal generator by initializing the student on a set of ODE trajectories obtained from the teacher. To achieve this, we adopt a hybrid forcing strategy that combines teacher forcing and diffusion forcing (mask shown on the right).
}
    \label{fig:hybrid_forcing}
\end{figure}

\subsection{Distillation Framework} \label{subsec:self_forcing}
We build our distillation framework on prior self-forcing~\citep{huang2025self}, an effective AR distillation paradigm designed to reduce exposure bias. Self-forcing simulates inference-time AR rollouts during training by predicting new chunks conditioned on the model’s own previously generated tokens (\ie, using generated history rather than ground-truth context). 

Many recent long-video distillation pipelines~\citep{liu2025rolling, yang2025longlive, cui2025self, lin2025apt2} train a long-horizon student model using a teacher that generates only short video segments (\eg, $5$-second windows from the Wan-2.1 teacher). However, this design decomposes long videos into loosely coupled short segments, limiting the student’s ability to retrieve and reason over long-range spatial memory. In contrast, we distill from a bidirectional teacher trained on much longer $20$-second clips, enabling a direct alignment between the student’s long-video distribution and the teacher’s long-video distribution.

\subsubsection{ODE Initialization with Hybrid Forcing}

Following recent AR video distillation frameworks~\citep{yin2025causvid, huang2025self}, we adopt the same ODE-initialization procedure to adapt the bidirectional teacher into a causal model and to enable the student to utilize long-horizon spatial memory. The student is initialized with the teacher’s weights and trained to regress precomputed ODE trajectories at the four denoising time steps used during distillation and inference.

Interestingly, we found that using teacher forcing or diffusion forcing alone produces suboptimal initialization for causal distillation. We therefore introduce an improved ODE initialization strategy that combines both paradigms to facilitate fast convergence. Specifically, given a training sequence of $B$ latent blocks, we divide them into two chunks: the first $B-K$ blocks contain clean, spatially compressed latents, while the remaining $K$ blocks contain uncompressed latents. For each block in the second chunk, we add noise at varying scales and condition the block causally on (i) past uncompressed latents within the second chunk, as in diffusion forcing, and (ii) the clean compressed latents from the first chunk, as in teacher forcing, as illustrated in~\cref{fig:hybrid_forcing}. This hybrid strategy provides a stronger warm-up for long-horizon memory retrieval compared to either forcing method alone.

\begin{figure}[t!]
    \centering
    \includegraphics[width=\textwidth]{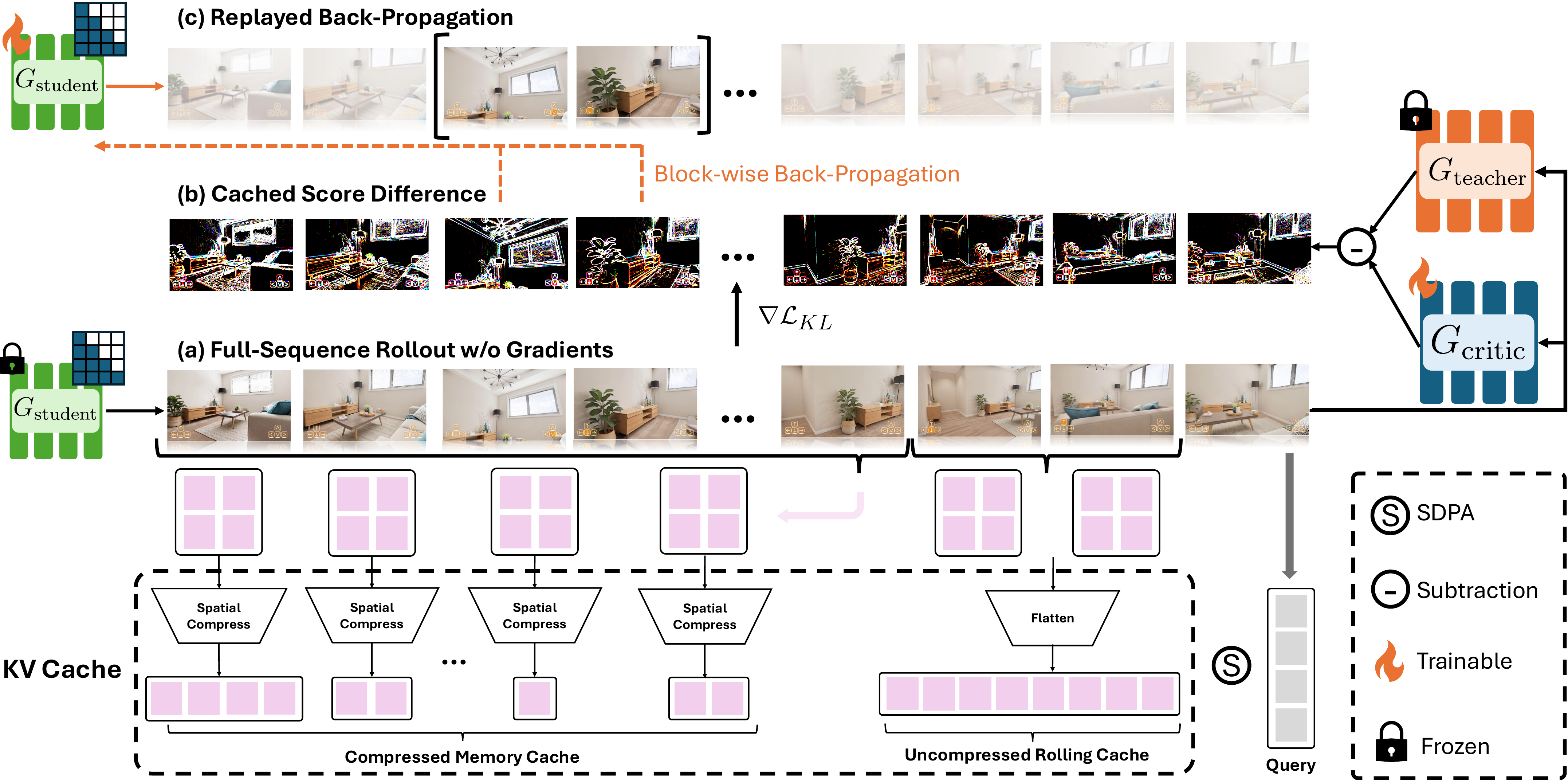}
    \caption{\textbf{Long video distillation with replayed back-propagation.} Given a 20-second bidirectional teacher, we distill it into a fast autoregressive student model via self-forcing~\citep{huang2025self}. We achieve memory-efficient distillation via replayed back-propagation. Specifically, (a) we first let the student model generate the full sequence via self-rollout with gradients disabled. (b) Then we compute and cache the DMD score difference maps over the entire predicted sequence using the critic and teacher models. (c) We re-run the student forward pass with auto-differentiation enabled and back-propagate the corresponding cached score difference maps to accumulate parameter gradients. Parameters are
updated once after the full replay.}
    \label{fig:dmd}
\end{figure}

\subsubsection{Long-Video Distillation with Replayed Back-propagation}

Self-forcing~\citep{huang2025self} employs the Distribution Matching Distillation (DMD) loss~\citep{yin2024dmd2, yin2024dmd1}, which minimizes the reverse KL divergence between the diffused data distribution and the student distribution across sampled timesteps~$u$. The gradient of the objective can be approximated by the difference between the real-data and generated-data score functions, $s^{\mathrm{data}}$ and $s^{\mathrm{gen}}$:

\begin{align}
\nabla_\theta \mathcal{L}_{KL}
\approx
-\mathbb{E}_{u}\!\left[
\int \bigl(
s^{\mathrm{data}}\!\left(\Psi(G_\theta(\epsilon, c_{\text{text}}), u\right)
-
s^{\mathrm{gen}}\!\left(\Psi(G_\theta(\epsilon, c_{\text{text}}), u\right)
\bigr)
\,\frac{d G_\theta(\epsilon, , c_{\text{text}})}{d\theta}\, d\epsilon
\right],
\end{align}

where $\Psi$ denotes the forward diffusion process, $\epsilon$ is Gaussian noise, and $G_\theta$ is the student generator.

The original self-forcing implementation becomes prohibitively memory-intensive for long-video distillation: the student must roll out an entire long video with autograd enabled, and the DMD loss must be back-propagated through the full computation graph. Since the graph grows linearly with video length, this quickly becomes intractable for long rollout scenarios.

To address this limitation, we introduce a \textit{replayed back-propagation} technique that stores only a small computation graph corresponding to a single generation block during self-rollout. This idea is related in spirit to prior work on neural rendering~\citep{zhang2022arf}, but our method incorporates substantial modifications to support distillation.
As shown in \cref{fig:dmd}, we first generate the entire predicted sequence of $L$ video latents using $G_\theta$ with autograd disabled,
\[
\hat{\mathbf{x}}_{0:L} = \text{stop-grad} (G_\theta(\epsilon_{0:L})),
\]
and compute the score-difference maps using frozen real and fine-tuned fake score models:
\[
\Delta \hat{s}_{0:L} = s^{\mathrm{data}} (\hat{\mathbf{x}}_{0:L}) - s^{\mathrm{gen}} (\hat{\mathbf{x}}_{0:L})
\]
Next, we replay the AR rollout block by block. For each block index $l$, we re-run the student AR forward pass with autograd enabled, conditioned on the previously generated context, and then back-propagate the corresponding cached score-difference map to update the gradients:

\begin{align}
\nabla_\theta \mathcal{L}_{KL} 
\approx 
\sum_{l=1}^{L} - \Delta \hat{s}_l
\frac{\partial G_{\theta}}{\partial \theta}.
\end{align}

After processing block $i$, its computation graph is immediately freed before moving to the next block. Parameters are updated once after the full replay. This approach shifts back-propagation from full-sequence differentiation to block-wise differentiation, reducing peak GPU memory from that of an entire rollout to that of a single video-latent block, while still capturing gradients that reflect the full-length video distribution of the teacher.

\subsection{Runtime Eff\text{i}ciency Optimization}
\label{subsec:runtime}

To provide users with a real-time experience, we optimize our codebase for efficient inference. Since inference latency and throughput are largely bounded by GPU memory bandwidth and CPU speed, we tailor our optimizations accordingly. We first apply \texttt{torch.compile} to reduce kernel launch overhead and memory costs, \eg, for RMSNorm, RoPE embeddings, and modulation layers. We also use a KV cache in self-attention to avoid recomputing historical context, and we store the cache in FP8 E4M3 format to halve memory usage and reduce memory transfer time during inference. We also employ FlashAttention v3~\citep{Shah2024FlashAttention3FA} with FP8 kernels to improve performance on NVIDIA Hopper GPUs. Finally, we manually fuse and reflow certain PyTorch operations based on profiling results to further reduce overhead.

After minimizing memory and CPU costs, we utilize parallelization to shard computation and memory loads across multiple GPUs. We adopt a parallelization strategy similar to that used in our long-sequence training (\cref{sec:implementation_details}); specifically, all linear layers and cross-attention modules are parallelized over the sequence dimension (sequence/context parallelism), while self-attention operators are parallelized over attention heads (tensor parallelism). We use NCCL \texttt{All-to-All} operations to switch tensor layouts between these two parallelization schemes. For example, when transitioning from sequence parallelism to tensor-parallel attention, an \texttt{All-to-All} operation scatters along the heads dimension and gathers along the sequence-length dimension simultaneously. We use tensor parallelism for self-attention because it enables sharding of the KV cache across devices, ensuring that each GPU stores only the heads it is responsible for computing.

\section{Experiments} 
\label{sec:experiment}

In this section, we detail our implementation setup, demonstrate the capabilities of \ours{}, and compare its performance against existing methods.

\subsection{Implementation details.}
\label{sec:implementation_details}

Training a 20-second 14B base model can be challenging as the model needs to process more than 120K tokens for a forward pass. This creates high pressure on GPU memory. To deal with this, we use a combination of existing techniques in both bidirectional long-horizon training and auto-regressive distillation stages: (i) employ Fully Sharded Data Parallel (FSDP)~\citep{zhao2023pytorch} to shard training batch,  model parameters, gradients and optimization states over GPUs; (ii) leverage sequence parallelism~\citep{li2023sequence} to scatter the sequence; and (iii) use tensor parallelism~\citep{shoeybi2019megatron} to distribute attention heads across GPUs. This linearly reduces the memory requirement by the number of available computes. 
For the student model, our memory spatial compression configuration is empirically set to $S=[1, 4, 2, 4, 4, 4, 2, 4, 4, 2, 4, 4, 4, 2, 4, 4, 2, 4]$ to fit a 20-second context into the original pre-trained token context length corresponding to a 5-second clip. We apply this configuration recurrently, where the compression ratio for latent frame $i$ is selected as  $s_{i}=S[i\ \text{mod}\ \text{len}(S)]$. 
During distillation, we progressively increase the rollout length to keep training stable and facilitate convergence. Specifically, we first roll out 5-second sequences for 250 training iterations using the intermediate 5-second teacher, then extend the rollout to 10 seconds with the 10-second teacher for another 150 iterations, and finally increase the horizon to 20 seconds with the 20-second teacher for the last 150 iterations. We further optimize decoding-time efficiency by replacing the original VAE with the same Tiny VAE used in MotionStream~\citep{shin2025motionstream}. Our final model is trained on 32 H100 GPUs, each with 80GB of memory.

\subsection{Capabilities Showcase}

\begin{figure}[h!]
 \centering
 \includegraphics[width=0.98\textwidth]{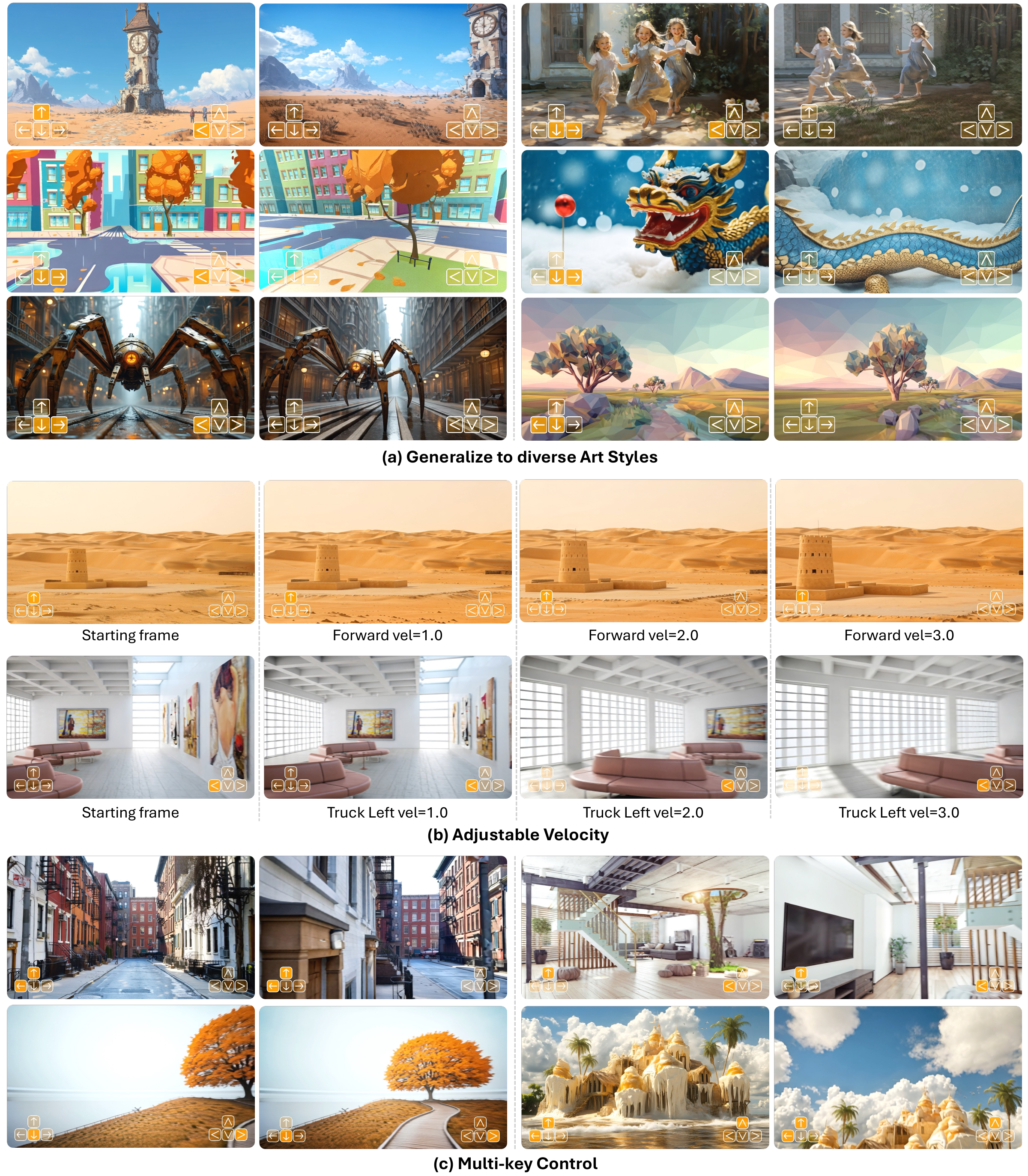}
 \caption{\ours{}~generates high-quality, diverse, and controllable videos while maintaining strong spatial consistency over long horizons. The figure showcases its generalization across varied artistic styles, its ability to follow adjustable camera velocities, and its support for complex multi-key control.}
 \label{fig:diversity}
\end{figure}

\ours{}~achieves high-quality, diverse, and precisely controllable video generation while maintaining strong spatial consistency over long horizons. \cref{fig:teaser}, \cref{fig:diversity}, and \cref{fig:qual_memory} illustrate these capabilities across a wide range of scenes and styles, and we refer readers to the videos on our project page for full results.

\paragraph{\textbf{Diversity.}}
\ours{} generalizes far beyond real indoor or outdoor environments. Starting from a single initial frame, it can explore worlds depicted in oil paintings, comic illustrations, vector art, low-poly renders, and various other stylized domains (\cref{fig:diversity}(a)). Notably, the model naturally exhibits correct distance awareness—faraway elements move more slowly than nearby objects—and demonstrates strong 3D shape understanding as the camera moves around subjects. This broad generalization capability enables \ours{} to be applied across a wide range of creative and visual settings.

\paragraph{\textbf{Long-horizon Memory.}}

Our \ours{} model enables robust spatial memory retrieval even under large camera movements, as shown in the first row of \cref{fig:teaser} and \cref{fig:qual_memory}. The model can accurately recover previously generated scene content with minimal loss of detail, even when that content has remained outside the camera’s field of view for an extended period. This is particularly noteworthy given that our model relies solely on compressed historical video tokens and contexts from absolute camera pose, without introducing any explicit 3D scene representation, handcrafted memory heuristics, or auxiliary hyper-networks.

\paragraph{\textbf{Adjustable Velocity.}}
Because camera actions are represented as continuous relative velocities, users can freely control the exploration speed by adjusting the displacement coefficient $\lambda$. As shown in \cref{fig:diversity}(b), \ours{} supports a wide spectrum of translational and rotational velocities while consistently producing high-quality, temporally stable outputs.

\paragraph{\textbf{Multi-Key Control.}}
\ours{}~responds reliably to complex multi-key inputs that combine both translation and rotation actions, enabling rich and intuitive interaction with the generated world. This high degree of motion freedom allows users to explore the scenes in real time with precision and flexibility.

\begin{table*}[t]
  \small
  \setlength{\tabcolsep}{6pt}
  \caption{
    \textbf{Quantitative comparison.} We compare \ours~with recent representative open-source world models on \textbf{20s video clips}. $^{\dagger}$We compute Subject Consistency, Background Consistency, Motion Smoothness, Dynamic Degree, Aesthetic Quality, Imaging Quality, and then calculate the average score over them.
  }
  % \vspace{0.3em}
  \label{tab:vbench}
  \centering
  \begin{tabular}{lcccccc}
      \toprule
      \multirow{2}{*}{Model} &
      \multicolumn{3}{c}{Visual quality $\uparrow$} &
      \multicolumn{2}{c}{Action accuracy (RPE $\downarrow$) } \\
      \cmidrule(lr){2-4}\cmidrule(lr){5-6}
       & Average Score$^{\dagger}$ & Image Quality & Aesthetic & Trans  & Rot \\
      \midrule
      Matrix-Game-2.0~\citep{he2025matrixgame2} & 0.7447	 & 0.6551	 &  0.4931	 & 0.1122 &  1.48 \\ 
      Hunyuan-GameCraft~\citep{li2025hunyuangame} & 0.7885	 & \textbf{0.6737} & 0.5874 & 0.1149 & 1.23  \\
      \midrule
      \ours ~(ours) & \textbf{0.8015} & 0.6665 & \textbf{0.5967} & \textbf{0.0906} & \textbf{1.00} \\
      \bottomrule
  \end{tabular}\\
\end{table*}

\subsection{Quantitative Comparison}
We construct a benchmark test set of 220 images sourced from Adobe Stock. The set spans both realistic scenes (landscapes, urban environments, indoor spaces) and non-realistic scenes (cartoons, vector art, oil paintings). 
The 220 images are randomly partitioned into 11 groups. 
For each group, we evaluate all baseline models using a predefined action script, resulting in 220 generated videos per baseline. 
The output duration is fixed at 20 seconds.
We further evaluate each baseline from two aspects: visual quality and action accuracy.
We compare with two state-of-the-art baselines: Matrix-Game-2.0~\citep{he2025matrixgame2} and Hunyuan-GameCraft~\citep{li2025hunyuangame}.

\paragraph{\textbf{Visual quality.}} 
We evaluate visual quality using selected dimensions from VBench~\citep{huang2023vbench}, with results summarized in \cref{tab:main}. \ours{} achieves the strongest overall performance among all compared baselines. Although trained at 480p resolution, it performs comparably to Hunyuan-GameCraft~\citep{li2025hunyuangame}, which is trained on 720P videos, in terms of Image Quality. Notably, it attains higher Aesthetics scores and significantly outperforms Matrix-Game 2.0~\citep{he2025matrixgame2} across metrics.

\paragraph{\textbf{Action Accuracy.}}
All baselines support control via the same set of discrete keyboard actions (W/A/S/D translation and camera rotation), although their internal mappings from actions to camera speed may differ. 
To ensure a fair comparison, we apply a behavioral evaluation protocol: every model executes the same predefined action sequence, and we reconstruct its induced camera trajectory from the generated video using ViPE~\citep{huang2025vipe}. 
The reconstructed trajectory is aligned to the canonical ground-truth camera trajectory using a Sim(3) Umeyama alignment~\citep{umeyama2002least}, which removes scale and coordinate-frame differences. 
We then evaluate translational and rotational Relative Pose Error (RPE-trans, RPE-rot) to measure how closely each model follows the intended camera trajectory, independent of internal speed tuning.
We show the results in \cref{tab:vbench}.
\ours~demonstrates the most faithful adherence to the target motion, resulting in the lowest overall RPE.

\begin{figure}[t]
    \centering
    \includegraphics[width=1\textwidth]{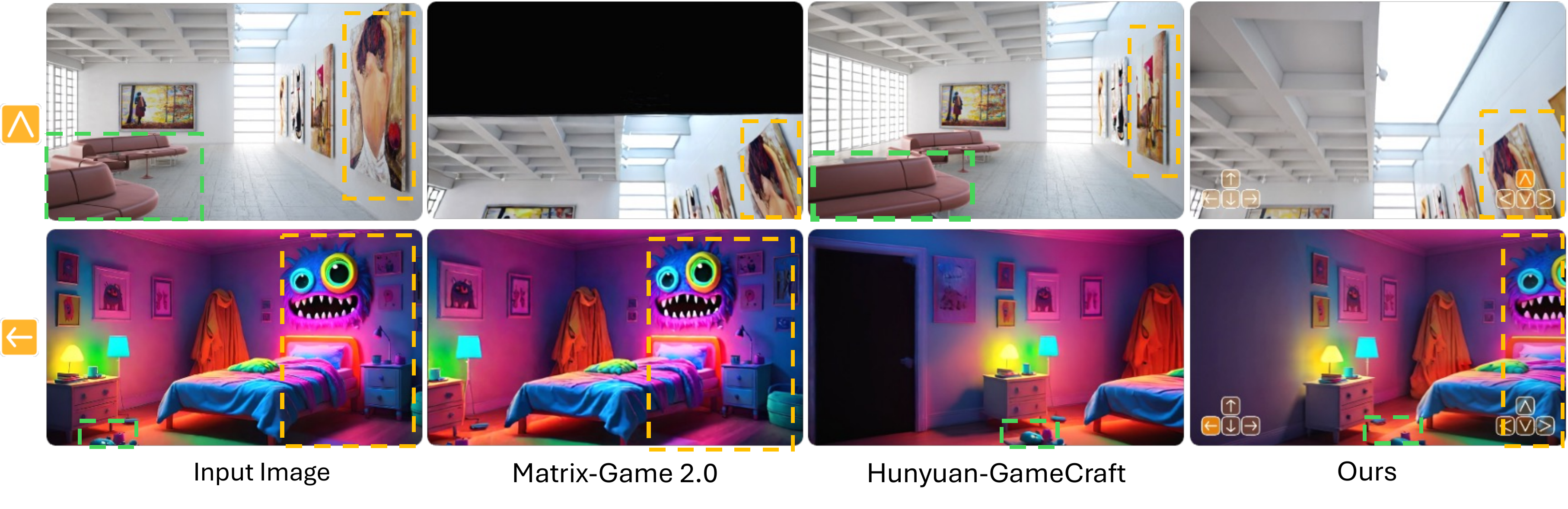}
    \caption{\textbf{Qualitative comparison of action control.}
    In the ``gallery'' example (top row), Hunyuan-GameCraft pans the camera left instead of rotating it upward (\ie \texttt{Tile Up}).
    Matrix-Game-2.0 nominally follows the intended action, but it introduces a large black region at the top of the frame.
    In the ``bedroom'' example (bottom row), 
    Hunyuan-GameCraft rotates the camera instead of translating left.
    Matrix-Game-2.0 drifts left instead of executing the commanded leftward rotation. 
    }
    \label{fig:qual_action}
\end{figure}

\subsection{Qualitative Comparison}
\paragraph{\textbf{Action accuracy.}}
We compare the accuracy of action control in \cref{fig:qual_action}. \ours\ adheres to the commanded actions most faithfully, producing motion that stays closest to the intended trajectory. For example, when applying \texttt{Tilt Up} $\bigwedge$, Matrix-Game-2.0~\citep{he2025matrixgame2} fails to generate new content at the top image boundary, resulting in a black void, while Hunyuan-GameCraft~\citep{li2025hunyuangame} exhibits negligible vertical camera movement. Similarly, when applying \texttt{tuck Left} $\leftarrow$, Hunyuan-GameCraft behaves more like \texttt{Pan Left} $<$, and Matrix-Game-2.0 incorrectly remains static instead of a lateral movement. In contrast, \ours\ accurately follows the trajectory, revealing the ceiling structure and shifting the viewing angle correctly without artifacts.

\paragraph{\textbf{Memory.}}

\begin{figure}[t]
    \centering
    \includegraphics[width=1.0\textwidth]{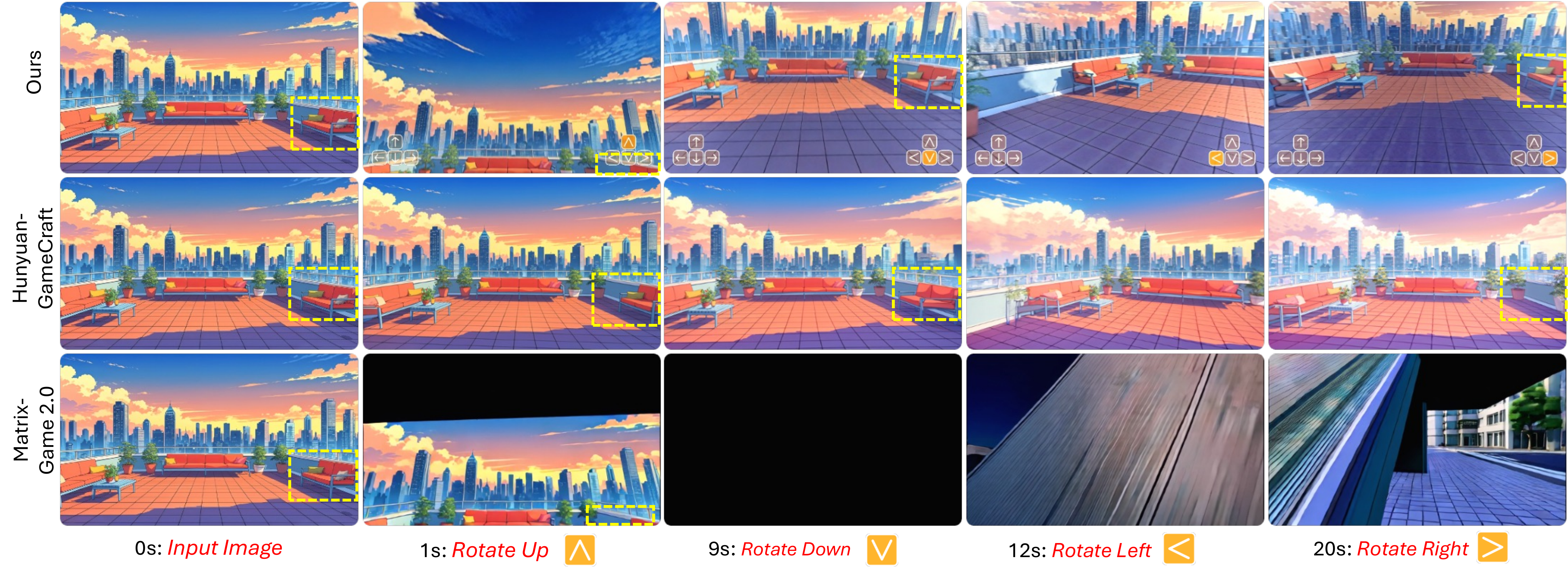}
    \caption{\textbf{Qualitative comparison of memory.} Previous methods, such as Hunyuan-GameCraft~\citep{li2025hunyuangame} and Matrix-Game-2.0~\citep{he2025matrixgame2}, forget the bench on the right-hand side in this case quickly.
    }
    \label{fig:qual_memory}
\end{figure}

Due to the difficulty in aligning all baselines with the same actions, 
we show a qualitative comparison on memory in \cref{fig:qual_memory}.
Previous baselines such as Hunyuan-GameCraft~\citep{li2025hunyuangame} and Matrix-Game-2.0~\citep{he2025matrixgame2} fail to maintain object persistence in this scenario. For example, Hunyuan-GameCraft forgets the bench once the viewpoint moves away and returns, and Matrix-Game 2.0 quickly loses the context of the input image, whereas our model consistently regenerates the previously observed content (also see \cref{fig:teaser}).

\begin{figure}[t]
    \centering
    \includegraphics[width=1.0\textwidth]
    {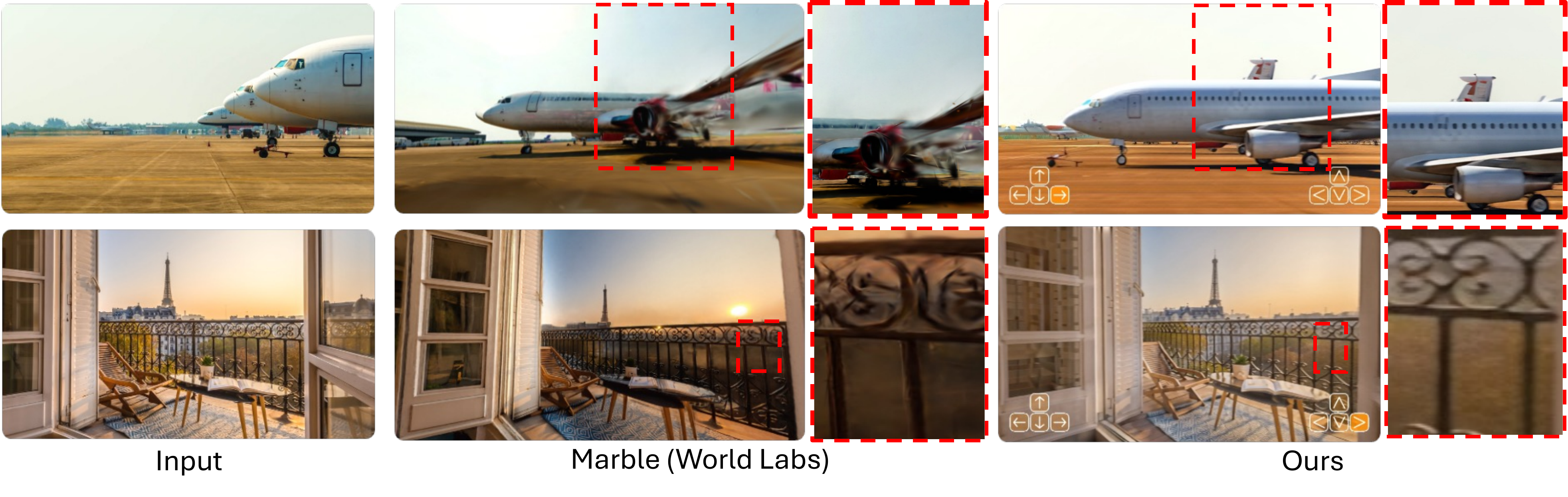}
    \caption{\textbf{Qualitative comparison with Marble from World Labs}~\citep{worldlabs}. We compare with a commercial solution, Marble.
    Since the final product from Marble is a Gaussian splatting~\citep{kerbl3Dgaussians} rendered result, it inevitably introduces artifacts such as Gaussian floaters.
    Our method \ours\ , instead, generates clean results.
    }
    \label{fig:qual_worldlabs}
\end{figure}

\paragraph{\textbf{Comparison with Marble.}}
We also qualitatively compare our method with Marble~\citep{worldlabs}, a commercial system whose output is based on Gaussian-splatting reconstruction~\citep{kerbl3Dgaussians}, in \cref{fig:qual_worldlabs}. 
Since Marble's final output is rendered frames from Gaussian Splatting, it inherently produces artifacts such as Gaussian floaters, which are visible in its results. In contrast, \ours~avoids these reconstruction artifacts and delivers clean, stable outputs.

\section{Discussion and Conclusion} 
\label{sec:conclusion}
\textbf{Limitations.} 
Our system still exhibits several limitations. First, the generated videos demonstrate limited diversity and restricted scene dynamics, primarily due to training on datasets composed mostly of static scenes rendered from the Unreal Engine. Additionally, our approach struggles to generate extremely long videos—on the scale of minutes. Moreover, the combination of large model size, KV cache requirements for long-horizon memory, and multiple iterative denoising steps significantly impacts inference latency under resource-constrained settings. Nonetheless, we believe these issues can be mitigated through targeted refinements to the pipeline and appropriate scaling of data and training.

\textbf{Conclusion.} 
In this work, we presented \ours, an interactive video world model that enables real-time inference and long-horizon spatial memory for virtual scene exploration. By integrating a lightweight, spatial-aware memory mechanism with the scalable Self-Forcing distillation paradigm, \ours\ enables consistent world generation from a single image, without relying on explicit geometric representations. Our method shows that integrating compressed historical latents with full-horizon supervision from a long-context teacher can successfully address challenges such as drifting and memory forgetting, which have long hindered video generation. The architectural innovations in \ours\ lay a scalable and adaptable foundation for general-purpose world simulators, with potential applications in embodied AI and immersive virtual content creation.

\clearpage
\newpage
\bibliographystyle{assets/plainnat}
\bibliography{paper}

@article{wu2025spatial,
  title={Video World Models with Long-term Spatial Memory},
  author={Wu, Tong and Yang, Shuai and Po, Ryan and Xu, Yinghao and Liu, Ziwei and Lin, Dahua and Wetzstein, Gordon},
  journal={arXiv preprint arXiv:2506.05284},
  year={2025}
}

@article{feng2024thematrix,
  title={The matrix: Infinite-horizon world generation with real-time moving control},
  author={Feng, Ruili and Zhang, Han and Yang, Zhantao and Xiao, Jie and Shu, Zhilei and Liu, Zhiheng and Zheng, Andy and Huang, Yukun and Liu, Yu and Zhang, Hongyang},
  journal={arXiv preprint arXiv:2412.03568},
  year={2024}
}

@inproceedings{zhang2022arf,
  title={Arf: Artistic radiance fields},
  author={Zhang, Kai and Kolkin, Nick and Bi, Sai and Luan, Fujun and Xu, Zexiang and Shechtman, Eli and Snavely, Noah},
  booktitle={European Conference on Computer Vision},
  pages={717--733},
  year={2022},
  organization={Springer}
}

@article{Shah2024FlashAttention3FA,
  title={FlashAttention-3: Fast and Accurate Attention with Asynchrony and Low-precision},
  author={Jay Shah and Ganesh Bikshandi and Ying Zhang and Vijay Thakkar and Pradeep Ramani and Tri Dao},
  journal={ArXiv},
  year={2024},
  volume={abs/2407.08608},
  url={https://api.semanticscholar.org/CorpusID:271098045}
}

@article{zhao2023pytorch,
  title={Pytorch fsdp: experiences on scaling fully sharded data parallel},
  author={Zhao, Yanli and Gu, Andrew and Varma, Rohan and Luo, Liang and Huang, Chien-Chin and Xu, Min and Wright, Less and Shojanazeri, Hamid and Ott, Myle and Shleifer, Sam and others},
  journal={arXiv preprint arXiv:2304.11277},
  year={2023}
}

@inproceedings{li2023sequence,
  title={Sequence parallelism: Long sequence training from system perspective},
  author={Li, Shenggui and Xue, Fuzhao and Baranwal, Chaitanya and Li, Yongbin and You, Yang},
  booktitle={Proceedings of the 61st Annual Meeting of the Association for Computational Linguistics (Volume 1: Long Papers)},
  pages={2391--2404},
  year={2023}
}

@article{shoeybi2019megatron,
  title={Megatron-lm: Training multi-billion parameter language models using model parallelism},
  author={Shoeybi, Mohammad and Patwary, Mostofa and Puri, Raul and LeGresley, Patrick and Casper, Jared and Catanzaro, Bryan},
  journal={arXiv preprint arXiv:1909.08053},
  year={2019}
}

@misc{bardes2023v,
  title={V-jepa: Latent video prediction for visual representation learning},
  author={Bardes, Adrien and Garrido, Quentin and Ponce, Jean and Chen, Xinlei and Rabbat, Michael and LeCun, Yann and Assran, Mido and Ballas, Nicolas},
  year={2023}
}

@article{hu2023gaia,
  title={Gaia-1: A generative world model for autonomous driving},
  author={Hu, Anthony and Russell, Lloyd and Yeo, Hudson and Murez, Zak and Fedoseev, George and Kendall, Alex and Shotton, Jamie and Corrado, Gianluca},
  journal={arXiv preprint arXiv:2309.17080},
  year={2023}
}

@article{tu2025drivingworldmodel,
      title={The Role of World Models in Shaping Autonomous Driving: A Comprehensive Survey}, 
      author={Tu, Sifan and Zhou, Xin and Liang, Dingkang and Jiang, Xingyu and Zhang, Yumeng and Li, Xiaofan and Bai, Xiang},
      journal={arXiv preprint arXiv:2502.10498},
      year={2025}
}

@inproceedings{zhou2025hermes,
  title={HERMES: A Unified Self-Driving World Model for Simultaneous 3D Scene Understanding and Generation},
  author={Zhou, Xin and Liang, Dingkang and Tu, Sifan and Chen, Xiwu and Ding, Yikang and Zhang, Dingyuan and Tan, Feiyang and Zhao, Hengshuang and Bai, Xiang},
  booktitle={Proceedings of the IEEE/CVF International Conference on Computer Vision},
  year={2025}
}

@article{ye2025yan,
  title={Yan: Foundational Interactive Video Generation},
  author={Ye, Deheng and Zhou, Fangyun and Lv, Jiacheng and Ma, Jianqi and Zhang, Jun and Lv, Junyan and Li, Junyou and Deng, Minwen and Yang, Mingyu and Fu, Qiang and others},
  journal={arXiv preprint arXiv:2508.08601},
  year={2025}
}

@article{he2025matrixgame2,
  title={Matrix-Game 2.0: An Open-Source, Real-Time, and Streaming Interactive World Model},
  author={He, Xianglong and Peng, Chunli and Liu, Zexiang and Wang, Boyang and Zhang, Yifan and Cui, Qi and Kang, Fei and Jiang, Biao and An, Mengyin and Ren, Yangyang and others},
  journal={arXiv preprint arXiv:2508.13009},
  year={2025}
}

@article{zhang2025matrixgame1,
  title={Matrix-Game: Interactive World Foundation Model},
  author={Zhang, Yifan and Peng, Chunli and Wang, Boyang and Wang, Puyi and Zhu, Qingcheng and Kang, Fei and Jiang, Biao and Gao, Zedong and Li, Eric and Liu, Yang and others},
  journal={arXiv preprint arXiv:2506.18701},
  year={2025}
}

@article{li2025hunyuangame,
  title={Hunyuan-GameCraft: High-dynamic Interactive Game Video Generation with Hybrid History Condition},
  author={Li, Jiaqi and Tang, Junshu and Xu, Zhiyong and Wu, Longhuang and Zhou, Yuan and Shao, Shuai and Yu, Tianbao and Cao, Zhiguo and Lu, Qinglin},
  journal={arXiv preprint arXiv:2506.17201},
  year={2025}
}

@misc{genie3,
  title         = {Genie 3: A New Frontier for World Models},
  author        = {Philip J. Ball and Jakob Bauer and Frank Belletti and Bethanie Brownfield and Ariel Ephrat and Shlomi Fruchter and Agrim Gupta and Kristian Holsheimer and Aleksander Holynski and Jiri Hron and Christos Kaplanis and Marjorie Limont and Matt McGill and Yanko Oliveira and Jack Parker-Holder and Frank Perbet and Guy Scully and Jeremy Shar and Stephen Spencer and Omer Tov and Ruben Villegas and Emma Wang and Jessica Yung and Cip Baetu and Jordi Berbel and David Bridson and Jake Bruce and Gavin Buttimore and Sarah Chakera and Bilva Chandra and Paul Collins and Alex Cullum and Bogdan Damoc and Vibha Dasagi and Maxime Gazeau and Charles Gbadamosi and Woohyun Han and Ed Hirst and Ashyana Kachra and Lucie Kerley and Kristian Kjems and Eva Knoepfel and Vika Koriakin and Jessica Lo and Cong Lu and Zeb Mehring and Alex Moufarek and Henna Nandwani and Valeria Oliveira and Fabio Pardo and Jane Park and Andrew Pierson and Ben Poole and Helen Ran and Tim Salimans and Manuel Sanchez and Igor Saprykin and Amy Shen and Sailesh Sidhwani and Duncan Smith and Joe Stanton and Hamish Tomlinson and Dimple Vijaykumar and Luyu Wang and Piers Wingfield and Nat Wong and Keyang Xu and Christopher Yew and Nick Young and Vadim Zubov and Douglas Eck and Dumitru Erhan and Koray Kavukcuoglu and Demis Hassabis and Zoubin Gharamani and Raia Hadsell and A{\"a}ron van den Oord and Inbar Mosseri and Adrian Bolton and Satinder Singh and Tim Rockt{\"a}schel},
  year          = {2025},
  url           = {https://deepmind.google/discover/blog/genie-3-a-new-frontier-for-world-models/}
}

@misc{magica2,
    author = {Lab, Dynamics},
    title = {Magica 2},
    url = {https://blog.dynamicslab.ai/},
    year = {2025},
}

@misc{worldlabs2025,
  title   = {Generating Worlds},
  author  = {{World Labs}},
  url     = {https://www.worldlabs.ai/blog/generating-worlds},
  note    = {Product site},
  year    = {2025}
}

@inproceedings{gu2025cosmos,
  title={Cosmos World Foundation Models for Physical AI},
  author={Gu, Jinwei},
  booktitle={Proceedings of the 3rd International Workshop on Rich Media With Generative AI},
  pages={39--39},
  year={2025}
}

@article{roth2025learned,
  title={Learned Perceptive Forward Dynamics Model for Safe and Platform-aware Robotic Navigation},
  author={Roth, Pascal and Frey, Jonas and Cadena, Cesar and Hutter, Marco},
  journal={arXiv preprint arXiv:2504.19322},
  year={2025}
}

@misc{rtfm2025,
  title   = {RTFM: A Real-Time Frame Model},
  author  = {{World Labs}},
  url     = {https://www.worldlabs.ai/blog/rtfm},
  year    = {2025}
}

@misc{videoworldsimulators2024,
  title={Video generation models as world simulators},
  author={Tim Brooks and Bill Peebles and Connor Holmes and Will DePue and Yufei Guo and Li Jing and David Schnurr and Joe Taylor and Troy Luhman and Eric Luhman and Clarence Ng and Ricky Wang and Aditya Ramesh},
  year={2024},
  url={https://openai.com/research/video-generation-models-as-world-simulators},
}

@misc{parkerholder2024genie2,
  title         = {Genie 2: A Large-Scale Foundation World Model},
  author        = {Jack Parker-Holder and Philip Ball and Jake Bruce and Vibhavari Dasagi and Kristian Holsheimer and Christos Kaplanis and Alexandre Moufarek and Guy Scully and Jeremy Shar and Jimmy Shi and Stephen Spencer and Jessica Yung and Michael Dennis and Sultan Kenjeyev and Shangbang Long and Vlad Mnih and Harris Chan and Maxime Gazeau and Bonnie Li and Fabio Pardo and Luyu Wang and Lei Zhang and Frederic Besse and Tim Harley and Anna Mitenkova and Jane Wang and Jeff Clune and Demis Hassabis and Raia Hadsell and Adrian Bolton and Satinder Singh and Tim Rockt{\"a}schel},
  year          = {2024},
  url           = {https://deepmind.google/discover/blog/genie-2-a-large-scale-foundation-world-model/}
}

@article{gao2025seedance,
  title={Seedance 1.0: Exploring the Boundaries of Video Generation Models},
  author={Gao, Yu and Guo, Haoyuan and Hoang, Tuyen and Huang, Weilin and Jiang, Lu and Kong, Fangyuan and Li, Huixia and Li, Jiashi and Li, Liang and Li, Xiaojie and others},
  journal={arXiv preprint arXiv:2506.09113},
  year={2025}
}

@article{lin2025apt2,
  title={Autoregressive Adversarial Post-Training for Real-Time Interactive Video Generation},
  author={Lin, Shanchuan and Yang, Ceyuan and He, Hao and Jiang, Jianwen and Ren, Yuxi and Xia, Xin and Zhao, Yang and Xiao, Xuefeng and Jiang, Lu},
  journal={arXiv preprint arXiv:2506.09350},
  year={2025}
}

@article{wiedemer2025veo,
  title={Video models are zero-shot learners and reasoners},
  author={Wiedemer, Thadd{\"a}us and Li, Yuxuan and Vicol, Paul and Gu, Shixiang Shane and Matarese, Nick and Swersky, Kevin and Kim, Been and Jaini, Priyank and Geirhos, Robert},
  journal={arXiv preprint arXiv:2509.20328},
  year={2025}
}

@article{wan2025wan,
  title={Wan: Open and advanced large-scale video generative models},
  author={Wan, Team and Wang, Ang and Ai, Baole and Wen, Bin and Mao, Chaojie and Xie, Chen-Wei and Chen, Di and Yu, Feiwu and Zhao, Haiming and Yang, Jianxiao and others},
  journal={arXiv preprint arXiv:2503.20314},
  year={2025}
}

@article{polyak2024moviegen,
  title={Movie gen: A cast of media foundation models},
  author={Polyak, Adam and Zohar, Amit and Brown, Andrew and Tjandra, Andros and Sinha, Animesh and Lee, Ann and Vyas, Apoorv and Shi, Bowen and Ma, Chih-Yao and Chuang, Ching-Yao and others},
  journal={arXiv preprint arXiv:2410.13720},
  year={2024}
}

@inproceedings{huang2025self,
  title={Self Forcing: Bridging the Train-Test Gap in Autoregressive Video Diffusion},
  author={Huang, Xun and Li, Zhengqi and He, Guande and Zhou, Mingyuan and Shechtman, Eli},
  booktitle={Advances in neural information processing systems},
  year={2025}
}

@inproceedings{yin2025causvid,
  title={From slow bidirectional to fast autoregressive video diffusion models},
  author={Yin, Tianwei and Zhang, Qiang and Zhang, Richard and Freeman, William T and Durand, Fredo and Shechtman, Eli and Huang, Xun},
  booktitle={Proceedings of the Computer Vision and Pattern Recognition Conference},
  pages={22963--22974},
  year={2025}
}

@article{teng2025magi,
  title={MAGI-1: Autoregressive Video Generation at Scale},
  author={Teng, Hansi and Jia, Hongyu and Sun, Lei and Li, Lingzhi and Li, Maolin and Tang, Mingqiu and Han, Shuai and Zhang, Tianning and Zhang, WQ and Luo, Weifeng and others},
  journal={arXiv preprint arXiv:2505.13211},
  year={2025}
}

@inproceedings{xie2025progressive,
  title={Progressive autoregressive video diffusion models},
  author={Xie, Desai and Xu, Zhan and Hong, Yicong and Tan, Hao and Liu, Difan and Liu, Feng and Kaufman, Arie and Zhou, Yang},
  booktitle={Proceedings of the Computer Vision and Pattern Recognition Conference},
  pages={6322--6332},
  year={2025}
}

@inproceedings{yin2024dmd1,
  title={One-step diffusion with distribution matching distillation},
  author={Yin, Tianwei and Gharbi, Micha{\"e}l and Zhang, Richard and Shechtman, Eli and Durand, Fredo and Freeman, William T and Park, Taesung},
  booktitle={Proceedings of the IEEE/CVF conference on computer vision and pattern recognition},
  pages={6613--6623},
  year={2024}
}

@article{shin2025motionstream,
  title={MotionStream: Real-Time Video Generation with Interactive Motion Controls},
  author={Shin, Joonghyuk and Li, Zhengqi and Zhang, Richard and Zhu, Jun-Yan and Park, Jaesik and Schechtman, Eli and Huang, Xun},
  journal={arXiv preprint arXiv:2511.01266},
  year={2025}
}

@article{Chung2023UniMaxFA,
  title={UniMax: Fairer and more Effective Language Sampling for Large-Scale Multilingual Pretraining},
  author={Hyung Won Chung and Noah Constant and Xavier Garc{\'i}a and Adam Roberts and Yi Tay and Sharan Narang and Orhan Firat},
  journal={ArXiv},
  year={2023},
  volume={abs/2304.09151},
  url={https://api.semanticscholar.org/CorpusID:258187051}
}

@article{yin2024dmd2,
  title={Improved distribution matching distillation for fast image synthesis},
  author={Yin, Tianwei and Gharbi, Micha{\"e}l and Park, Taesung and Zhang, Richard and Shechtman, Eli and Durand, Fredo and Freeman, Bill},
  journal={Advances in neural information processing systems},
  volume={37},
  pages={47455--47487},
  year={2024}
}

@inproceedings{henschel2025streamingt2v,
  title={Streamingt2v: Consistent, dynamic, and extendable long video generation from text},
  author={Henschel, Roberto and Khachatryan, Levon and Poghosyan, Hayk and Hayrapetyan, Daniil and Tadevosyan, Vahram and Wang, Zhangyang and Navasardyan, Shant and Shi, Humphrey},
  booktitle={Proceedings of the Computer Vision and Pattern Recognition Conference},
  pages={2568--2577},
  year={2025}
}

@article{chen2025skyreels2,
  title={Skyreels-v2: Infinite-length film generative model},
  author={Chen, Guibin and Lin, Dixuan and Yang, Jiangping and Lin, Chunze and Zhu, Junchen and Fan, Mingyuan and Zhang, Hao and Chen, Sheng and Chen, Zheng and Ma, Chengcheng and others},
  journal={arXiv preprint arXiv:2504.13074},
  year={2025}
}

@misc{song2023consistency,
  title={Consistency models},
  author={Song, Yang and Dhariwal, Prafulla and Chen, Mark and Sutskever, Ilya},
  year={2023}
}

@article{geng2025meanflow,
  title={Mean flows for one-step generative modeling},
  author={Geng, Zhengyang and Deng, Mingyang and Bai, Xingjian and Kolter, J Zico and He, Kaiming},
  journal={arXiv preprint arXiv:2505.13447},
  year={2025}
}

@article{yang2025longlive,
      title={LongLive: Real-time Interactive Long Video Generation}, 
      author={Shuai Yang and Wei Huang and Ruihang Chu and Yicheng Xiao and Yuyang Zhao and Xianbang Wang and Muyang Li and Enze Xie and Yingcong Chen and Yao Lu and Song Han and Yukang Chen},
      year={2025},
      journal={arXiv preprint arXiv:2509.22622},
}

@article{xiao2025worldmem,
  title={Worldmem: Long-term consistent world simulation with memory},
  author={Xiao, Zeqi and Lan, Yushi and Zhou, Yifan and Ouyang, Wenqi and Yang, Shuai and Zeng, Yanhong and Pan, Xingang},
  journal={arXiv preprint arXiv:2504.12369},
  year={2025}
}

@article{ha2018worldmodels,
  title={World models},
  author={Ha, David and Schmidhuber, J{\"u}rgen},
  journal={arXiv preprint arXiv:1803.10122},
  volume={2},
  number={3},
  year={2018}
}

@article{huang2025voyager,
  title={Voyager: Long-Range and World-Consistent Video Diffusion for Explorable 3D Scene Generation},
  author={Huang, Tianyu and Zheng, Wangguandong and Wang, Tengfei and Liu, Yuhao and Wang, Zhenwei and Wu, Junta and Jiang, Jie and Li, Hui and Lau, Rynson WH and Zuo, Wangmeng and Guo, Chunchao},
  journal={arXiv preprint arXiv:2506.04225},
  year={2025}
}

@article{team2025hunyuanworld,
  title={Hunyuanworld 1.0: Generating immersive, explorable, and interactive 3d worlds from words or pixels},
  author={Team, HunyuanWorld and Wang, Zhenwei and Liu, Yuhao and Wu, Junta and Gu, Zixiao and Wang, Haoyuan and Zuo, Xuhui and Huang, Tianyu and Li, Wenhuan and Zhang, Sheng and others},
  journal={arXiv preprint arXiv:2507.21809},
  year={2025}
}

@article{kong2024hunyuanvideo,
  title={Hunyuanvideo: A systematic framework for large video generative models},
  author={Kong, Weijie and Tian, Qi and Zhang, Zijian and Min, Rox and Dai, Zuozhuo and Zhou, Jin and Xiong, Jiangfeng and Li, Xin and Wu, Bo and Zhang, Jianwei and others},
  journal={arXiv preprint arXiv:2412.03603},
  year={2024}
}

@article{po2025long,
  title={Long-context state-space video world models},
  author={Po, Ryan and Nitzan, Yotam and Zhang, Richard and Chen, Berlin and Dao, Tri and Shechtman, Eli and Wetzstein, Gordon and Huang, Xun},
  journal={arXiv preprint arXiv:2505.20171},
  year={2025}
}

@article{song2025history,
  title={History-guided video diffusion},
  author={Song, Kiwhan and Chen, Boyuan and Simchowitz, Max and Du, Yilun and Tedrake, Russ and Sitzmann, Vincent},
  journal={arXiv preprint arXiv:2502.06764},
  year={2025}
}

@article{zhang2025lact,
  title={Test-time training done right},
  author={Zhang, Tianyuan and Bi, Sai and Hong, Yicong and Zhang, Kai and Luan, Fujun and Yang, Songlin and Sunkavalli, Kalyan and Freeman, William T and Tan, Hao},
  journal={arXiv preprint arXiv:2505.23884},
  year={2025}
}

@article{wang2025spatialvid,
  title={SpatialVID: A Large-Scale Video Dataset with Spatial Annotations},
  author={Wang, Jiahao and Yuan, Yufeng and Zheng, Rujie and Lin, Youtian and Gao, Jian and Chen, Lin-Zhuo and Bao, Yajie and Zhang, Yi and Zeng, Chang and Zhou, Yanxi and others},
  journal={arXiv preprint arXiv:2509.09676},
  year={2025}
}

@article{li2025sekai,
  title={Sekai: A Video Dataset towards World Exploration},
  author={Li, Zhen and Li, Chuanhao and Mao, Xiaofeng and Lin, Shaoheng and Li, Ming and Zhao, Shitian and Xu, Zhaopan and Li, Xinyue and Feng, Yukang and Sun, Jianwen and others},
  journal={arXiv preprint arXiv:2506.15675},
  year={2025}
}

@article{che2024gamegenx,
  title={Gamegen-x: Interactive open-world game video generation},
  author={Che, Haoxuan and He, Xuanhua and Liu, Quande and Jin, Cheng and Chen, Hao},
  journal={arXiv preprint arXiv:2411.00769},
  year={2024}
}

@article{yu2025gamefactory,
  title={Gamefactory: Creating new games with generative interactive videos},
  author={Yu, Jiwen and Qin, Yiran and Wang, Xintao and Wan, Pengfei and Zhang, Di and Liu, Xihui},
  journal={arXiv preprint arXiv:2501.08325},
  year={2025}
}

@inproceedings{anderson2018vln,
  title={Vision-and-language navigation: Interpreting visually-grounded navigation instructions in real environments},
  author={Anderson, Peter and Wu, Qi and Teney, Damien and Bruce, Jake and Johnson, Mark and S{\"u}nderhauf, Niko and Reid, Ian and Gould, Stephen and Van Den Hengel, Anton},
  booktitle={Proceedings of the IEEE conference on computer vision and pattern recognition},
  pages={3674--3683},
  year={2018}
}

@article{hong2020vlnbert,
  title={A recurrent vision-and-language bert for navigation},
  author={Hong, Yicong and Wu, Qi and Qi, Yuankai and Rodriguez-Opazo, Cristian and Gould, Stephen},
  journal={arXiv preprint arXiv:2011.13922},
  year={2020}
}

@article{mao2025yume,
  title={Yume: An Interactive World Generation Model},
  author={Mao, Xiaofeng and Lin, Shaoheng and Li, Zhen and Li, Chuanhao and Peng, Wenshuo and He, Tong and Pang, Jiangmiao and Chi, Mingmin and Qiao, Yu and Zhang, Kaipeng},
  journal={arXiv preprint arXiv:2507.17744},
  year={2025}
}

@inproceedings{zhang2025world,
  title={World-consistent video diffusion with explicit 3d modeling},
  author={Zhang, Qihang and Zhai, Shuangfei and Martin, Miguel Angel Bautista and Miao, Kevin and Toshev, Alexander and Susskind, Joshua and Gu, Jiatao},
  booktitle={Proceedings of the Computer Vision and Pattern Recognition Conference},
  pages={21685--21695},
  year={2025}
}

@article{huang2025vid2world,
  title={Vid2World: Crafting Video Diffusion Models to Interactive World Models},
  author={Huang, Siqiao and Wu, Jialong and Zhou, Qixing and Miao, Shangchen and Long, Mingsheng},
  journal={arXiv preprint arXiv:2505.14357},
  year={2025}
}

@article{yu2025contextasmem,
  title={Context as memory: Scene-consistent interactive long video generation with memory retrieval},
  author={Yu, Jiwen and Bai, Jianhong and Qin, Yiran and Liu, Quande and Wang, Xintao and Wan, Pengfei and Zhang, Di and Liu, Xihui},
  journal={arXiv preprint arXiv:2506.03141},
  year={2025}
}

@article{wu2025geometry,
  title={Geometry forcing: Marrying video diffusion and 3d representation for consistent world modeling},
  author={Wu, Haoyu and Wu, Diankun and He, Tianyu and Guo, Junliang and Ye, Yang and Duan, Yueqi and Bian, Jiang},
  journal={arXiv preprint arXiv:2507.07982},
  year={2025}
}

@article{chen2025deepverse4d,
      title={DeepVerse: 4D Autoregressive Video Generation as a World Model}, 
      author={Junyi Chen and Haoyi Zhu and Xianglong He and Yifan Wang and Jianjun Zhou and Wenzheng Chang and Yang Zhou and Zizun Li and Zhoujie Fu and Jiangmiao Pang and Tong He},
      year={2025},
      journal={arXiv preprint arXiv:2506.01103},
}

@article{zhang2025framepack,
      title={Packing Input Frame Context in Next-Frame Prediction Models for Video Generation}, 
      author={Lvmin Zhang and Maneesh Agrawala},
      year={2025},
      journal={arXiv preprint arXiv:2504.12626},
}

@article{kodaira2025streamdit,
      title={StreamDiT: Real-Time Streaming Text-to-Video Generation}, 
      author={Akio Kodaira and Tingbo Hou and Ji Hou and Masayoshi Tomizuka and Yue Zhao},
      year={2025},
      journal={arXiv preprint arXiv:2507.03745},
}

@inproceedings{ma2025yousee,
  title={You see it, you got it: Learning 3d creation on pose-free videos at scale},
  author={Ma, Baorui and Gao, Huachen and Deng, Haoge and Luo, Zhengxiong and Huang, Tiejun and Tang, Lulu and Wang, Xinlong},
  booktitle={Proceedings of the Computer Vision and Pattern Recognition Conference},
  pages={2016--2029},
  year={2025}
}

@inproceedings{ren2025gen3c,
  title={Gen3c: 3d-informed world-consistent video generation with precise camera control},
  author={Ren, Xuanchi and Shen, Tianchang and Huang, Jiahui and Ling, Huan and Lu, Yifan and Nimier-David, Merlin and M{\"u}ller, Thomas and Keller, Alexander and Fidler, Sanja and Gao, Jun},
  booktitle={Proceedings of the Computer Vision and Pattern Recognition Conference},
  pages={6121--6132},
  year={2025}
}

@inproceedings{yu2025wonderworld,
  title={Wonderworld: Interactive 3d scene generation from a single image},
  author={Yu, Hong-Xing and Duan, Haoyi and Herrmann, Charles and Freeman, William T and Wu, Jiajun},
  booktitle={Proceedings of the Computer Vision and Pattern Recognition Conference},
  pages={5916--5926},
  year={2025}
}

@article{kim2024fifo,
  title={Fifo-diffusion: Generating infinite videos from text without training},
  author={Kim, Jihwan and Kang, Junoh and Choi, Jinyoung and Han, Bohyung},
  journal={Advances in Neural Information Processing Systems},
  volume={37},
  pages={89834--89868},
  year={2024}
}

@inproceedings{sun2025ardiffuse,
  title={Ar-diffusion: Asynchronous video generation with auto-regressive diffusion},
  author={Sun, Mingzhen and Wang, Weining and Li, Gen and Liu, Jiawei and Sun, Jiahui and Feng, Wanquan and Lao, Shanshan and Zhou, SiYu and He, Qian and Liu, Jing},
  booktitle={Proceedings of the Computer Vision and Pattern Recognition Conference},
  pages={7364--7373},
  year={2025}
}

@article{chen2024dforcing,
  title={Diffusion forcing: Next-token prediction meets full-sequence diffusion},
  author={Chen, Boyuan and Mart{\'\i} Mons{\'o}, Diego and Du, Yilun and Simchowitz, Max and Tedrake, Russ and Sitzmann, Vincent},
  journal={Advances in Neural Information Processing Systems},
  volume={37},
  pages={24081--24125},
  year={2024}
}

@article{ruhe2024rollingdiff,
      title={Rolling Diffusion Models}, 
      author={David Ruhe and Jonathan Heek and Tim Salimans and Emiel Hoogeboom},
      year={2024},
      journal={arXiv preprint arXiv:2402.09470},
}

@article{liu2025rollingforcing,
      title={Rolling Forcing: Autoregressive Long Video Diffusion in Real Time}, 
      author={Kunhao Liu and Wenbo Hu and Jiale Xu and Ying Shan and Shijian Lu},
      year={2025},
      journal={arXiv preprint arXiv:2509.25161}, 
}

@article{alonso2024atari,
  title={Diffusion for world modeling: Visual details matter in atari},
  author={Alonso, Eloi and Jelley, Adam and Micheli, Vincent and Kanervisto, Anssi and Storkey, Amos J and Pearce, Tim and Fleuret, Fran{\c{c}}ois},
  journal={Advances in Neural Information Processing Systems},
  volume={37},
  pages={58757--58791},
  year={2024}
}

@article{jin2024pyramidal,
  title={Pyramidal flow matching for efficient video generative modeling},
  author={Jin, Yang and Sun, Zhicheng and Li, Ningyuan and Xu, Kun and Jiang, Hao and Zhuang, Nan and Huang, Quzhe and Song, Yang and Mu, Yadong and Lin, Zhouchen},
  journal={arXiv preprint arXiv:2410.05954},
  year={2024}
}

@article{valevski2024gamengen,
  title={Diffusion models are real-time game engines},
  author={Valevski, Dani and Leviathan, Yaniv and Arar, Moab and Fruchter, Shlomi},
  journal={arXiv preprint arXiv:2408.14837},
  year={2024}
}

@inproceedings{dalal2025one,
  title={One-minute video generation with test-time training},
  author={Dalal, Karan and Koceja, Daniel and Xu, Jiarui and Zhao, Yue and Han, Shihao and Cheung, Ka Chun and Kautz, Jan and Choi, Yejin and Sun, Yu and Wang, Xiaolong},
  booktitle={Proceedings of the Computer Vision and Pattern Recognition Conference},
  pages={17702--17711},
  year={2025}
}

@inproceedings{zhou2025tamingtf,
  title={Taming teacher forcing for masked autoregressive video generation},
  author={Zhou, Deyu and Sun, Quan and Peng, Yuang and Yan, Kun and Dong, Runpei and Wang, Duomin and Ge, Zheng and Duan, Nan and Zhang, Xiangyu},
  booktitle={Proceedings of the Computer Vision and Pattern Recognition Conference},
  pages={7374--7384},
  year={2025}
}

@article{gao2024ca2,
  title={Ca2-vdm: Efficient autoregressive video diffusion model with causal generation and cache sharing},
  author={Gao, Kaifeng and Shi, Jiaxin and Zhang, Hanwang and Wang, Chunping and Xiao, Jun and Chen, Long},
  journal={arXiv preprint arXiv:2411.16375},
  year={2024}
}

@article{hu2024acdit,
  title={Acdit: Interpolating autoregressive conditional modeling and diffusion transformer},
  author={Hu, Jinyi and Hu, Shengding and Song, Yuxuan and Huang, Yufei and Wang, Mingxuan and Zhou, Hao and Liu, Zhiyuan and Ma, Wei-Ying and Sun, Maosong},
  journal={arXiv preprint arXiv:2412.07720},
  year={2024}
}

@article{wang2025error,
  title={Error analyses of auto-regressive video diffusion models: A unified framework},
  author={Wang, Jing and Zhang, Fengzhuo and Li, Xiaoli and Tan, Vincent YF and Pang, Tianyu and Du, Chao and Sun, Aixin and Yang, Zhuoran},
  journal={arXiv preprint arXiv:2503.10704},
  year={2025}
}

@article{cui2025self,
  title={Self-Forcing++: Towards Minute-Scale High-Quality Video Generation},
  author={Cui, Justin and Wu, Jie and Li, Ming and Yang, Tao and Li, Xiaojie and Wang, Rui and Bai, Andrew and Ban, Yuanhao and Hsieh, Cho-Jui},
  journal={arXiv preprint arXiv:2510.02283},
  year={2025}
}

@article{liu2025rolling,
  title={Rolling Forcing: Autoregressive Long Video Diffusion in Real Time},
  author={Liu, Kunhao and Hu, Wenbo and Xu, Jiale and Shan, Ying and Lu, Shijian},
  journal={arXiv preprint arXiv:2509.25161},
  year={2025}
}

@article{kurai2025magiccraft,
  title={MagicCraft: Natural Language-Driven Generation of Dynamic and Interactive 3D Objects for Commercial Metaverse Platforms},
  author={Kurai, Ryutaro and Hiraki, Takefumi and Hiroi, Yuichi and Hirao, Yutaro and Perusqu{\'\i}a-Hern{\'a}ndez, Monica and Uchiyama, Hideaki and Kiyokawa, Kiyoshi},
  journal={arXiv preprint arXiv:2504.21332},
  year={2025}
}

@article{lu2025dreamart,
  title={Dreamart: Generating interactable articulated objects from a single image},
  author={Lu, Ruijie and Liu, Yu and Tang, Jiaxiang and Ni, Junfeng and Wang, Yuxiang and Wan, Diwen and Zeng, Gang and Chen, Yixin and Huang, Siyuan},
  journal={arXiv preprint arXiv:2507.05763},
  year={2025}
}

@inproceedings{xia2025drawer,
  title={Drawer: Digital reconstruction and articulation with environment realism},
  author={Xia, Hongchi and Su, Entong and Memmel, Marius and Jain, Arhan and Yu, Raymond and Mbiziwo-Tiapo, Numfor and Farhadi, Ali and Gupta, Abhishek and Wang, Shenlong and Ma, Wei-Chiu},
  booktitle={Proceedings of the Computer Vision and Pattern Recognition Conference},
  pages={21771--21782},
  year={2025}
}

@inproceedings{luo2025physpart,
  title={Physpart: Physically plausible part completion for interactable objects},
  author={Luo, Rundong and Geng, Haoran and Deng, Congyue and Li, Puhao and Wang, Zan and Jia, Baoxiong and Guibas, Leonidas and Huang, Siyuan},
  booktitle={2025 IEEE International Conference on Robotics and Automation (ICRA)},
  pages={12386--12393},
  year={2025},
  organization={IEEE}
}

@inproceedings{kreber2025guiding,
  title={Guiding diffusion-based articulated object generation by partial point cloud alignment and physical plausibility constraints},
  author={Kreber, Jens U and Stueckler, Joerg},
  booktitle={Proceedings of the IEEE/CVF International Conference on Computer Vision},
  pages={3206--3214},
  year={2025}
}

@article{peng2023yarn,
  title={Yarn: Efficient context window extension of large language models},
  author={Peng, Bowen and Quesnelle, Jeffrey and Fan, Honglu and Shippole, Enrico},
  journal={arXiv preprint arXiv:2309.00071},
  year={2023}
}

@article{su2024rope,
  title={Roformer: Enhanced transformer with rotary position embedding},
  author={Su, Jianlin and Ahmed, Murtadha and Lu, Yu and Pan, Shengfeng and Bo, Wen and Liu, Yunfeng},
  journal={Neurocomputing},
  volume={568},
  pages={127063},
  year={2024},
  publisher={Elsevier}
}

@article{yuan2025tarsier2,
  title={Tarsier2: Advancing large vision-language models from detailed video description to comprehensive video understanding},
  author={Yuan, Liping and Wang, Jiawei and Sun, Haomiao and Zhang, Yuchen and Lin, Yuan},
  journal={arXiv preprint arXiv:2501.07888},
  year={2025}
}

@inproceedings{chen2024internvl,
  title={Internvl: Scaling up vision foundation models and aligning for generic visual-linguistic tasks},
  author={Chen, Zhe and Wu, Jiannan and Wang, Wenhai and Su, Weijie and Chen, Guo and Xing, Sen and Zhong, Muyan and Zhang, Qinglong and Zhu, Xizhou and Lu, Lewei and others},
  booktitle={Proceedings of the IEEE/CVF conference on computer vision and pattern recognition},
  pages={24185--24198},
  year={2024}
}

@article{bai2023qwen,
  title={Qwen technical report},
  author={Bai, Jinze and Bai, Shuai and Chu, Yunfei and Cui, Zeyu and Dang, Kai and Deng, Xiaodong and Fan, Yang and Ge, Wenbin and Han, Yu and Huang, Fei and others},
  journal={arXiv preprint arXiv:2309.16609},
  year={2023}
}

@misc{openai_gpt5_2024,
  author = {OpenAI},
  title = {GPT-5.1},
  year = {2024},
  url = {https://www.openai.com},
  note = {Large language model}
}

@InProceedings{huang2023vbench,
     title={{VBench}: Comprehensive Benchmark Suite for Video Generative Models},
     author={Huang, Ziqi and He, Yinan and Yu, Jiashuo and Zhang, Fan and Si, Chenyang and Jiang, Yuming and Zhang, Yuanhan and Wu, Tianxing and Jin, Qingyang and Chanpaisit, Nattapol and Wang, Yaohui and Chen, Xinyuan and Wang, Limin and Lin, Dahua and Qiao, Yu and Liu, Ziwei},
     booktitle={Proceedings of the IEEE/CVF Conference on Computer Vision and Pattern Recognition},
     year={2024}
 }

@inproceedings{huang2025vipe,
        title={ViPE: Video Pose Engine for 3D Geometric Perception},
        author={Huang, Jiahui and Zhou, Qunjie and Rabeti, Hesam and Korovko, Aleksandr and Ling, Huan and Ren, Xuanchi and Shen, Tianchang and Gao, Jun and Slepichev, Dmitry and Lin, Chen-Hsuan and Ren, Jiawei and Xie, Kevin and Biswas, Joydeep and Leal-Taixe, Laura and Fidler, Sanja},
        booktitle={NVIDIA Research Whitepapers},
        year={2025}
    }

@misc{worldlabs,
  title   = {Marble},
  author  = {{World Labs}},
  url     = {https://marble.worldlabs.ai/},
  note    = {Product site},
  urldate = {2025-10-23},
  year    = {2025}
}

@Article{kerbl3Dgaussians,
      author       = {Kerbl, Bernhard and Kopanas, Georgios and Leimk{\"u}hler, Thomas and Drettakis, George},
      title        = {3D Gaussian Splatting for Real-Time Radiance Field Rendering},
      journal      = {ACM Transactions on Graphics},
      number       = {4},
      volume       = {42},
      month        = {July},
      year         = {2023},
      url          = {https://repo-sam.inria.fr/fungraph/3d-gaussian-splatting/}
}

@article{umeyama2002least,
  title={Least-squares estimation of transformation parameters between two point patterns},
  author={Umeyama, Shinji},
  journal={IEEE Transactions on pattern analysis and machine intelligence},
  volume={13},
  number={4},
  pages={376--380},
  year={2002},
  publisher={IEEE}
}

\end{document}